# Asimovian Adaptive Agents

**Diana F. Gordon**                                    GORDON@AIC.NRL.NAVY.MIL
*Navy Center for Applied Research in Artificial Intelligence*
*Naval Research Laboratory, Code 5515*
*Washington, D.C. 20375-5337 USA*

## Abstract

The goal of this research is to develop agents that are adaptive *and* predictable *and* timely. At first blush, these three requirements seem contradictory. For example, adaptation risks introducing undesirable side effects, thereby making agents' behavior less predictable. Furthermore, although formal verification can assist in ensuring behavioral predictability, it is known to be time-consuming.

Our solution to the challenge of satisfying all three requirements is the following. Agents have finite-state automaton plans, which are adapted online via evolutionary learning (perturbation) operators. To ensure that critical behavioral constraints are always satisfied, agents' plans are first formally verified. They are then *re*verified after every adaptation. If reverification concludes that constraints are violated, the plans are repaired. The main objective of this paper is to improve the efficiency of reverification after learning, so that agents have a sufficiently rapid response time. We present two solutions: positive results that certain learning operators are a priori guaranteed to preserve useful classes of behavioral assurance constraints (which implies that no reverification is needed for these operators), and efficient incremental reverification algorithms for those learning operators that have negative a priori results.

## 1. Introduction

Agents are becoming increasingly prevalent and effective. Robots and softbots, working individually or in concert, can relieve people of a great deal of labor-intensive tedium in their jobs as well as in their day-to-day lives. Designers can furnish agents with plans to perform desired tasks. Nevertheless, a designer cannot possibly foresee all circumstances that will be encountered by the agent. Therefore, in addition to supplying an agent with plans, it is essential to also enable the agent to learn and modify its plans to adapt to unforeseen circumstances. The introduction of learning, however, often makes the agent's behavior significantly harder to predict.[1] The goal of this research is to verify the behavior of adaptive agents. In particular, our objective is to develop efficient methods for determining whether the behavior of learning agents remains within the bounds of prespecified constraints (called "properties") after learning. This includes verifying that properties are preserved for single adaptive agents as well as verifying that global properties are preserved for multiagent systems in which one or more agents may adapt.

An example of a property is Asimov's First Law (Asimov, 1950). This law, which has also been studied by Weld and Etzioni (1994), states that an agent may not harm a human or allow a human to come to harm. The main contribution of Weld and Etzioni is a " 'call to arms:' before we release autonomous agents into real-world environments, we need some credible and computationally tractable means of making them obey Asimov's First Law...how do we stop our artifacts from causing us harm in the process of obeying our orders?" Of course, this law is too general for direct implementation and needs to be operationalized into specific properties testable on a system, such as "Never delete a user's file." This paper addresses Weld and Etzioni's call to arms in the context of adaptive agents. To respond to the call to arms, we are working toward "Asimovian" adaptive agents, which we define to be adaptive agents that can verify, in a reasonably efficient manner, whether user-defined properties are preserved after adaptation.[2] Such agents will either constrain their adaptation methods, or repair themselves in such a way as to preserve these properties.

---

[1] Even adding a simple, elegant learning mechanism such as chunking in Soar can substantially reduce system predictability (Soar project members, personal communication).

The verification method assumed here, *model checking*, consists of building a finite model of a system and checking whether the desired property holds in that model. In the context of this paper, model checking determines whether $S \models P$ for plan $S$ and property $P$, i.e., whether plan $S$ "models" (satisfies) property $P$. The output is either "yes" or "no" and, if "no," one or more counterexamples are provided. Model checking has proven to be very effective for safety-critical applications, e.g., a model checker uncovered a potentially disastrous error in a system designed to make buildings more earthquake resistant. This error would have unleashed a structural force to worsen earthquake vibrations, rather than dampen them (Elseaidy et al., 1994).

Essentially, model checking is brute force search through the set of all reachable states of the plan to check if the property holds. If the plan has a finite number of states, this process terminates. Model checking global properties of a multiagent plan has time complexity that is exponential in the number of agents.[3] With a large number of agents, this is could be a serious problem. In fact, even model checking a single agent plan with a huge number of states can be computationally prohibitive. A great deal of research in the verification community is currently focused on reduction techniques for handling very large state spaces (Clarke & Wing, 1997). One of the largest systems model checked to date using these reduction techniques had $10^{120}$ states (Burch et al., 1994). Nevertheless, the applicability of many of these reduction techniques is restricted and few are completely automated. Furthermore, none of them are tailored for efficient *reverification* after learning has altered the system. Some methods in the literature are designed for software that changes. One that emphasizes efficiency, as ours does, is Sokolsky and Smolka's (1994). However none of them, including Sokolsky and Smolka's method, are applicable to multiagent systems in which a single agent could adapt, thereby altering the global behavior of the overall system. In contrast, our approach addresses the timeliness of adaptive multiagent systems.

Consider how reverification fits into our overall adaptive agents framework. In this framework (see Figure 1), there are one or more agents with "anytime" plans (Grefenstette & Ramsey, 1992), i.e., plans that are continually executed in response to internal and external environmental conditions. Each agent's plan is assumed to be in the form of a finite-state automaton (FSA). FSAs have been shown to be effective representations of reactive agent plans/strategies (Burkhard, 1993; Kabanza, 1995; Carmel & Markovitch, 1996; Fogel, 1996).

---

2. They are also called *APT agents* because they are adaptive, predictable and timely.
3. The states in a multiagent plan are formed by taking the Cartesian product of states in the individual agent plans (see Section 3).

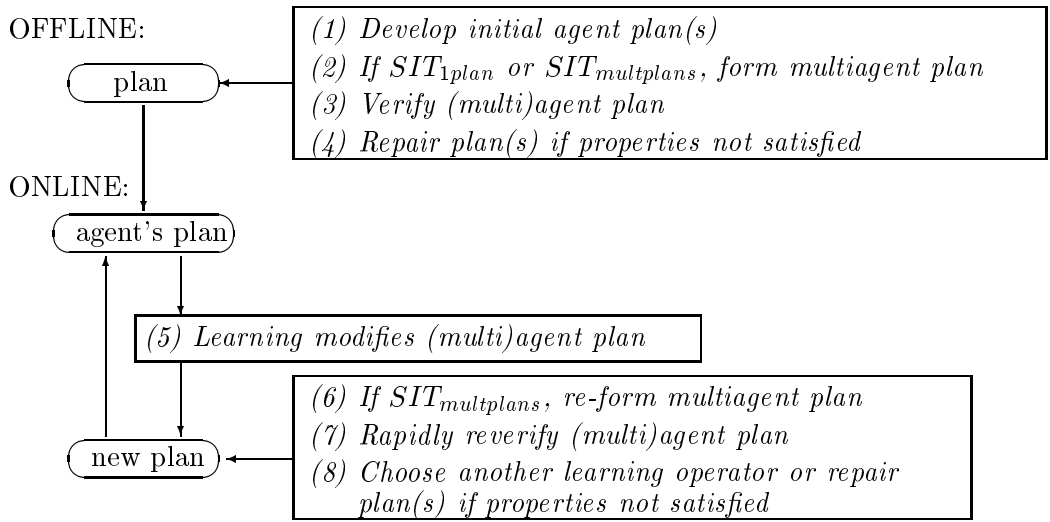

Figure 1: Verifiable adaptive agents.

Let us begin with step 1 in Figure 1. There are at least a couple of ways that the FSA plans could be formed initially. For one, a human plan designer could engineer the initial plans. This may require considerable effort and knowledge. An appealing alternative is to evolve (i.e., learn using evolutionary algorithms) the initial plans in a simulated environment. Fogel (1996) outlines a procedure for evolving FSAs that is effective for a number of problems, including an iterated version of the Prisoner's Dilemma.

Human plan engineers or evolutionary algorithms can develop plans that satisfy an agent's goals to a high degree. However, to provide strict behavioral guarantees, formal verification is also required. Therefore we assume that prior to fielding the agents, the (multi)agent plan has been verified offline to determine whether it satisfies critical properties (steps 2 and 3). If not, the plan is repaired (step 4). Plan repair is not addressed in this paper, although it is an important topic for future research. Steps 2 through 4 require some clarification. If there is a single agent, then it has one FSA plan and that is all that is verified and repaired, if needed. We call this $SIT_{1agent}$. (This notation, as well as other notation used in the paper, is included in the glossary of Appendix A.) If there are multiple agents that cooperate, we consider two possibilities. In $SIT_{1plan}$, every agent uses the same multiagent plan, which is a "product" of the individual agent plans. This multiagent plan is formed and verified to see if it satisfies global multiagent coordination properties. The multiagent plan is repaired if verification produces any errors, i.e., failure of the plan to satisfy a property. In $SIT_{multplans}$, each agent independently uses its own individual plan. To verify global properties, one of the agents takes the product of these individual plans to form a multiagent plan. This multiagent plan is what is verified. For $SIT_{multplans}$, one or more individual plans are repaired if the property is not satisfied.

After the initial plan(s) have been verified and repaired, the agents are fielded. While fielded (online), the agents apply learning (e.g., evolutionary operators) to their plan(s) as needed (step 5). Learning may be required to adapt the plan to handle unexpected situations or to fine-tune the plan. If $SIT_{1agent}$ or $SIT_{1plan}$, the single (multi)agent plan is adapted. If $SIT_{multplans}$, an agent adapts its own FSA, after which the multiagent (product) plan is *re*-formed. For all situations, one agent then rapidly *re*verifies the new (multi)agent plan to ensure it still satisfies the required properties (steps 6 and 7). Reformation of the multiagent plan and reverification are required to be as time-efficient as possible because they are performed online, perhaps in a highly time-critical situation. Whenever (re)verification fails, it produces a counterexample that is used to guide the choice of an alternative learning operator or other plan repair as needed (step 8). This process of executing, adapting, and reverifying plans cycles indefinitely as needed. The main focus of this paper is steps 6 and 7.

Rapid reverification after learning is a key to achieving timely agent responses. Our longterm goal is to examine all learning methods and important property classes to determine the quickest reverification method for each combination of learning method and property class. In this paper we present new results that certain useful learning operators are a priori guaranteed to be "safe" with respect to important classes of properties. In other words, if the property holds for the plan prior to learning, then it is guaranteed to still hold after learning.[4] If an agent uses these learning operators, it will be guaranteed to preserve the properties with *no re*verification required, i.e., steps 6 through 8 in Figure 1 need not be executed. This is the best one could hope for in an online situation where rapid response time is critical. For other learning operators and property classes our a priori results are negative. However, for the cases in which we have negative results, we present novel *incremental* reverification algorithms. These methods localize the reverification in order to save time over total reverification from scratch.[5] We also present a novel algorithm for efficiently re-forming a multiagent plan, for the situation ($SIT_{multplans}$) in which there are multiple agents, each learning independently.

The novelty of our approach is not in machine learning or verification per se, but rather the synthesis of the two. There are numerous important potential applications of our approach. For example, if antiviruses evolve more effective behaviors to combat viruses, we need to ensure that they do not evolve undesirable virus-like behavior. Another example is data mining agents that can flexibly adapt their plans to dynamic computing environments but whose behavior is adequately constrained for operation within secure or proprietary domains. A third example is planetary rovers that adapt to unforeseen conditions while remaining within critical mission parameters. Yet another example is automated factories that adapt to equipment failures but continue operation within essential tolerances and other specifications. Also, there are ongoing discussions at the Universities Space Research Association about launching orbiting unmanned vehicles to run laboratory experiments. The experiments would be semiautomated, and would thus require both adaptation and behavioral assurances.

The last important application that we will mention is in the domain of power grid and telecommunications networks. The following is an event that occurred (*The New York Times*, September 21, 1991, Business Section). In 1991 in New York, local electric utilities had a demand overload. In attempting to assist in solving the regional shortfall, AT&T put its own generators on the local power grid. This was a manual adaptation, but such adaptations are expected to become increasingly automated in the future. As a result of AT&T's actions, there was a local power overload and AT&T lost its own power, which resulted in a breakdown of the AT&T regional communications network. The regional network breakdown propagated to create a national breakdown in communications systems. This breakdown also triggered failures of many other control networks across the country, such as the air traffic control network. Air travel nationwide was shut down. In the future, it is reasonable to expect that some network controllers will be implemented using multiple, distributed cooperating software agents. This example dramatically illustrates the potential vulnerability of our national resources unless these agents satisfy *all* of the following criteria: continuous execution/monitoring, flexible adaptation to failures, safety/reliability, and timely responses. Our approach ensures that agents satisfy all of these.

---

4. This idea of property-preserving learning transformations was first introduced by Gordon (1998).

5. Incremental methods are often used in computer science for improving the time-efficiency of software.

This paper is organized as follows. Section 2 provides an illustrative example that is used throughout the paper. Section 3 has the necessary background definitions of FSAs, property types, formal verification, and machine learning operators. A priori results for specific machine learning operators are in Section 4. These learning operators alter automaton edges and the *transition conditions* associated with edges. A transition condition specifies the condition under which a state-to-state transition may be made. We present positive a priori results for some of these operators, where a "positive a priori result" means that the learning operator preserves a specified class of properties. On the other hand, counterexamples are presented to show that some of the learning operators do not necessarily preserve these properties. Section 5 extends the a priori results for the multiagent situation $SIT_{multiplans}$.

For all cases where we obtain negative a priori results, Section 6 provides incremental algorithms for re-forming the multiagent plan and reverifying it, along with a worst-case complexity analysis and empirical time complexity results. The empirical results show as much as a $\frac{1}{2}$-billion-fold speedup for one of the incremental algorithms over standard verification. The paper concludes with a discussion of related work and ideas for future research.

## 2. Illustrative Example

We begin with a multiagent example for $SIT_{1plan}$ or $SIT_{multiplans}$ that is used throughout the paper to illustrate the definitions and ideas. The section starts by addressing $SIT_{multiplans}$, where multiple agents have their own independent plans. Later in the section we address $SIT_{1plan}$, where each agent uses a joint multiagent plan.

Imagine a scenario where a vehicle has landed on a planet for the purpose of exploration and sample collection, for example as in the Pathfinder mission to Mars. Like the Pathfinder, there is a lander (called agent "L") from which a mobile rover emerges. However, in this case there are two rovers: the far ("F") rover for distant exploration, and the intermediary ("I") rover for transferring data and samples from F to L.

We assume an agent designer has developed the initial plans for F, I, and L, shown in Figures 2 and 3. These are simplified, rather than realistic, plans – for the purpose of illustration. Basically, rover F is either collecting samples/data (in state COLLECTING) or it is delivering them to rover I (when F is in its state DELIVERING). Rover I can either be receiving samples/data from rover F (when I is in its RECEIVING state) or it can deliver them to lander L (when it is in its DELIVERING state). If L is in its RECEIVING state, then it can receive the samples/data from I. Otherwise, L could be busy transmitting data to Earth (in state TRANSMITTING) or pausing between actions (in state PAUSING).

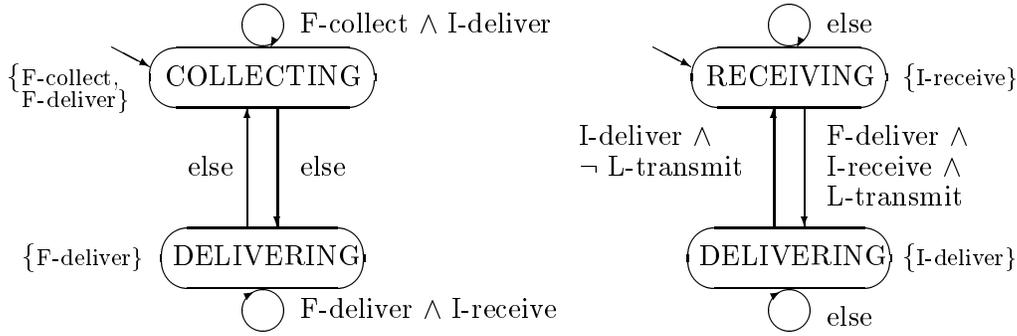

Figure 2: Plans for rovers F (left) and I (right).

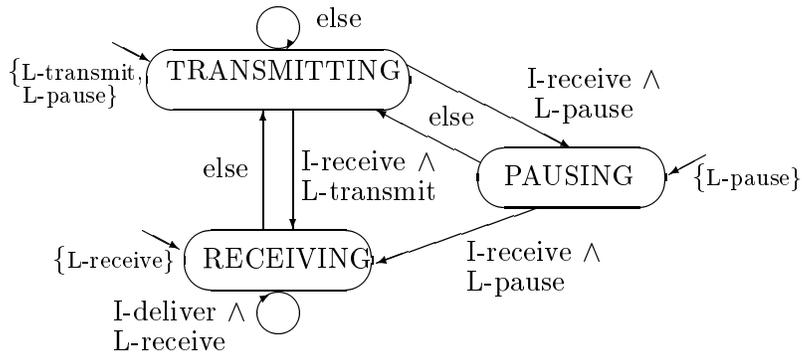

Figure 3: Plan for the lander L.

As mentioned above, plans are represented using FSAs. An FSA has a finite set of states (i.e., the vertices) and allowable state-to-state transitions (i.e., the directed edges between vertices). The purpose of having states is to divide the agent's overall task into subtasks. A state with an incoming arrow not from any other state is an *initial state*. Plan execution begins in an initial state.

Plan execution occurs as the agent takes actions, such as agent F taking action F-collect or F-deliver. Each agent has a repertoire of possible actions, a subset of which may be taken from each of its states. A plan designer can specify this subset for each state. The choice of a particular action from this subset is modeled in the FSA as nondeterministic. It is assumed that further criteria, not specified here, are used to make the final run-time choice of a single action from a state.

Let us specify the set of actions for each of the agents (F, I, L) in our example. F has two possible actions: F-collect and F-deliver. The first action means that F collects samples and/or data, and the second action means that it delivers these items to I. Rover I also has two actions: I-receive and I-deliver. The first action means I receives samples/data from F, and the second means that it delivers these items to L. L has three actions: L-transmit, L-pause, and L-receive. The first action means L transmits data to Earth, the second that it pauses between operations, and the third that it receives samples/data from I. For each FSA, the set of allowable actions from each state is specified in Figures 2 and 3 in small font next to the state. For example, rover F can only take action F-deliver from its DELIVERING state.

The *transition conditions* (i.e., the logical expressions labeling the edges) in an FSA plan describe the set of actions that enable a state-to-state transition to occur. The operator ∧ means "AND," ∨ means "OR," and ¬ means "NOT." The condition "else" will be defined shortly. The transition conditions of one agent can refer to the actions of one or more other agents. This is because each agent is assumed to be reactive to what it has observed other agents doing. If not visible, agents communicate their action choice.

Once an agent's action repertoire and its allowable actions from each state have been defined, "else" can be defined. The transition condition "else" labeling an outgoing edge from a state is an abbreviation denoting the set of all remaining actions that may be taken from the state that are not already covered by other transition conditions. For example, in Figure 3, L's three transition conditions from state TRANSMITTING are (I-receive ∧ L-transmit), (I-receive ∧ L-pause), and "else." L can only take L-transmit or L-pause from this state. However, rover I could take I-deliver instead of I-receive. Therefore, in this case "else" is equivalent to ((I-deliver ∧ L-transmit) ∨ (I-deliver ∧ L-pause)).

An FSA plan represents a set of allowable action sequences. In particular, a plan is the set of all action sequences that begin in an initial state and obey the transition conditions. An example action sequence allowed by F's plan is ((F-collect ∧ I-deliver), (F-collect ∧ I-receive), (F-deliver ∧ I-receive), ...) where F takes its actions and observes I's actions at each step in the sequence.

At run-time, these FSA plans are interpreted in the following manner. At every discrete time step, every agent (F, I, L) is at one of the states in its plan, and it selects the next action to take. Agents choose their actions independently. They do not need to synchronize on action choice. The choice of action might be based, for example, on sensory inputs from the environment. Although a complete plan would include the basis for action choice, as mentioned above, here we leave it unspecified in the FSA plans. Our rationale for doing this is that that the focus of this paper is on the verification of properties about correct action sequences. The basis for action choice is irrelevant to these properties.

Once each agent has chosen an action, all agents are assumed to observe the actions of the other agents that are mentioned in its FSA transition conditions. For example, F's transition conditions mention I's actions, so F needs to observe what I did. Based on its own action and those of the other relevant agent(s), an agent knows the next state to which it will transition. There is only one possible next state because the FSAs are assumed to be deterministic. For example, if F is in its COLLECTING state, and it chooses action F-collect, and it observes I taking action I-deliver, then it will stay in its COLLECTING state. The process of being in a state, choosing an action, observing the actions of other agents, then moving to a next state, is repeated indefinitely.

So far, we have been assuming $SIT_{multplans}$ where each agent has its own individual plan. If we assume $SIT_{1plan}$, then each agent uses the same multiagent plan to decide its actions. A multiagent plan is formed by taking a "product" (defined in Subsection 3.1) of the plans for F, I, and L. This product models the synchronous behavior of the agents, where "synchronous" means that at each time step every agent takes an action, observes actions of other agents, and then transitions to a next state. The product plan is formed, essentially, by taking the Cartesian product of the individual automaton states and the in-tersection of the transition conditions. Multiagent actions enable state-to-state transitions in the product plan. For example, if the agents jointly take the actions F-deliver and I-receive and L-transmit, then all agents will transition from the joint state (COLLECTING, RECEIVING, TRANSMITTING) to the joint state (DELIVERING, DELIVERING, RECEIVING) represented by triples of states in the FSAs for F, I, and L. A multiagent plan consists of the set of all action sequences that begin in a joint initial state of the product plan and obey the transition conditions.

Whether the situation is $SIT_{multplans}$ or $SIT_{1plan}$, a multiagent plan needs to be formed to verify global multiagent coordination properties (see step 2 of Figure 1). Verification of global properties consists of asking whether *all* of the action sequences allowed by the product plan satisfy the property.

One class of (global) properties of particular importance, which is addressed here, is that of forbidden multiagent actions that we want our agents to always avoid, called *Invariance* properties. An example is property P1: ¬(I-deliver ∧ L-transmit), which states that it should always be the case that I does not deliver at the same time that L is transmitting. This property prevents problems that may arise from the lander simultaneously receiving new data from I while transmitting older data to Earth. The second important class addressed here is *Response* properties. These properties state that if a particular multiagent action (the "trigger") has occurred, then eventually another multiagent action (the necessary "response") will occur. An example is property P2: If F-deliver has occurred, then eventually L will execute L-receive.

If the plans in Figures 2 and 3 are combined into a multiagent plan, will this multiagent plan satisfy properties P1 and P2? Answering this question is probably difficult or impossible for most readers if the determination is based on visual inspection of the FSAs. Yet there are only a couple of very small, simple FSAs in this example! This illustrates how even a few simple agents, when interacting, can exhibit complex global behaviors, thereby making global agent behavior difficult to predict. Clearly there is a need for rigorous behavioral guarantees, especially as the number and complexity of agents increases. Model checking fully automates this process. According to our model checker, the product plan for F, I, and L satisfies properties P1 and P2.

Rigorous guarantees are also needed after learning. Suppose lander L's transmitter gets damaged. Then one learning operator that could be applied is to delete L's action L-transmit, which thereafter prevents this action from being taken from state TRANSMITTING. After applying a learning operator, reverification may be required. For this particular operator (deleting an action), no reverification is needed (see Section 4).

In a multiagent situation, what gets modified by learning? Who forms and verifies the product FSA? And who performs repairs if verification fails, and what is repaired? The answers to these questions depend on whether it is $SIT_{1plan}$ or $SIT_{multplans}$. If $SIT_{1plan}$, the agent with the greatest computational power, e.g., lander L in our example, maintains the product plan by applying learning to it, verifying it, repairing it as needed, and then sending a copy of it to all of the agents to use. If $SIT_{multplans}$, an agent applies learning to its own individual plan. The individual plans are then sent to the computationally powerful agent, who forms the product and verifies that properties are satisfied. If repairs are needed, one or more agents repair their own individual plans.

It is assumed here that machine learning operators are applied one-at-a-time per agent rather than in batch and, if $SIT_{multplans}$, the agents co-evolve plans by taking turns learning (Potter, 1997). Beyond these assumptions, this paper does not focus on the learning operators per se (other than to define them). It focuses instead on the outcome resulting from the application of a learning operator. In particular, we address the reverification issue. The next section gives useful background definitions needed for understanding reverification.

## 3. Preliminary Definitions

This section provides definitions of FSAs, properties, verification, and machine learning operators. For a clear, unambiguous understanding of the results in this paper, many of these definitions are formal.

### 3.1 Automata for Agents' Plans

FSAs have at least four advantages over classical plans (Nilsson, 1980; Dean & Wellman, 1991). For one, unlike classical plans, the type of finite-state automaton plans used here allows potentially infinite (indeterminate) length action sequences.[6] This provides a good model of embedded agents that are continually responsive to their environment without any artificial termination to their behavior. Execution and learning may be interleaved in a natural manner. Another advantage is that FSA plans have states, and the plan designer can use these states to represent subtasks of the overall task. This subdivides the plan into smaller units, thereby potentially increasing the comprehensibility of the plans. States also enable different action choices at different times, even if the sensory inputs are the same. A third advantage of FSA plans is that they are particularly well-suited to modeling the concurrent behavior of multiple agents. An arbitrary number of single-agent plans can be developed independently and then composed into a synchronous multiagent plan (for which global properties may be tested) in a straightforward manner. Finally, FSA plans can be verified using the very popular and effective *automata-theoretic* model checking methods, e.g., see Kurshan (1994).

A disadvantage of FSA plans as opposed to classical plans is that there is a great deal of research that has been done on automatically forming classical plans, e.g., see Dean and Wellman (1991). It is unclear how much of this might be applicable to FSAs. On the other hand, evolutionary algorithms can be used to evolve FSA plans (Fogel, 1996). A disadvantage of FSA plans as opposed to plans composed of rule sets is that the latter may express a plan more succinctly. Nevertheless for plans that require formal verification, FSAs are preferable because the complex interactions that can occur between rules make them very hard to verify. Formal verification for FSAs is quite sophisticated and widely used in safety-critical industrial applications.

This subsection, which is based on Kurshan (1994), briefly summarizes the basics of the FSAs used to model agent plans. Figures 2 and 3 illustrate the definitions. This paper focuses on FSAs that model agents with a potentially infinite lifetime, represented as an infinite-length "string" (i.e., a sequence of actions).

---

6. Results for agents with finite lifetimes may be found in Gordon (1998, 1999).

Before beginning our discussion of automata, we briefly digress to define Boolean algebra. Examples throughout this paper have automaton transition conditions expressed in Boolean algebra, because Boolean algebra succinctly summarizes these transition conditions. Boolean algebra is also useful for succinctly expressing the properties. Furthermore, it is easier for us to describe two of the incremental reverification algorithms if we use Boolean algebra notation. Therefore, we briefly summarize the basics of Boolean algebra.

A Boolean algebra $\mathcal{K}$ is a set of elements with distinguished elements 0 and 1, closed under the Boolean $\wedge$, $\vee$, and $\neg$ operations, and satisfying the standard properties (Sikorski, 1969). For elements $x$ and $y$ of $\mathcal{K}$, $x \wedge y$ is called the *meet* of $x$ and $y$, $x \vee y$ is called the *join* of $x$ and $y$, and $\neg x$ is called the *complement* of $x$. For those readers who are unfamiliar with Boolean algebras and who want some intuition for these operations, it may help to imagine that each element of $\mathcal{K}$ is itself a set, e.g., a set of actions. Meet, join, and complement would then be set intersection, union, and complement, respectively. Elements 0 and 1, in this case, would be the empty set ($\emptyset$) and the set of all elements in the universe ($U$), respectively.

The Boolean algebras are assumed to be finite. There is a partial order among the elements, $\preceq$, which is defined as $x \preceq y$ if and only if $x \wedge y = x$. It may help to think of $\preceq$ as analogous to $\subseteq$ for sets. The elements 0 and 1 are defined as $\forall x \in \mathcal{K}$, $0 \preceq x$, and $\forall x \in \mathcal{K}$, $x \preceq 1$. The *atoms* (analogous to single-element sets) of $\mathcal{K}$, $\Gamma(\mathcal{K})$, are the nonzero elements of $\mathcal{K}$ minimal with respect to $\preceq$. In the rovers example, agents F, I, and L each have their own Boolean algebra with its atoms. The atoms of F's Boolean algebra are its actions F-collect and F-deliver; the atoms of I's algebra are I-receive and I-deliver; the atoms of L's algebra are L-transmit, L-pause, and L-receive. The element (F-collect $\vee$ F-deliver) of F's Boolean algebra describes the set of actions {F-collect, F-deliver}.

A Boolean algebra $\mathcal{K}_i$ is a *subalgebra* of $\mathcal{K}$ if $\mathcal{K}_i$ is a nonempty subset of $\mathcal{K}$ that is closed under the operations $\wedge$, $\vee$, and $\neg$, and also has the distinguished elements 0 and 1. $\prod \mathcal{K}_i$ is the *product algebra* of subalgebras $\mathcal{K}_i$. An atom of the product algebra is the meet of the atoms of the subalgebras. For example, if $a_1, ..., a_n$ are atoms of subalgebras $\mathcal{K}_1, ..., \mathcal{K}_n$, respectively, then $a_1 \wedge ... \wedge a_n$ is an atom of $\prod_{i=1}^{n} \mathcal{K}_i$.

The Boolean algebra $\mathcal{K}_F$ for agent F's actions is the smallest one containing the atoms of F's algebra. It contains all Boolean elements formed from F's atoms using the Boolean operators $\wedge$, $\vee$, and $\neg$, including 0 and 1. These same definitions hold for I and L's algebras $\mathcal{K}_I$ and $\mathcal{K}_L$. $\mathcal{K}_F \mathcal{K}_I \mathcal{K}_L$ is the product algebra used for all transition conditions in the multiagent plan (i.e., the product of the F, I, and L FSAs). One atom of the product algebra $\mathcal{K}_F \mathcal{K}_I \mathcal{K}_L$ is (F-collect $\wedge$ I-receive $\wedge$ L-pause). This is the form of actions taken simultaneously by the three agents. Algebras $\mathcal{K}_F$, $\mathcal{K}_I$, and $\mathcal{K}_L$ are subalgebras of the product algebra $\mathcal{K}_F \mathcal{K}_I \mathcal{K}_L$.

Let us return now to automata. Formally, an FSA of the type considered here is a three-tuple $S = (V(S), M_\mathcal{K}(S), I(S))$ where $V(S)$ is the set of vertices (states) of $S$, $\mathcal{K}$ is the Boolean algebra corresponding to $S$, $M_\mathcal{K}(S) : V(S) \times V(S) \to \mathcal{K}$ is the matrix of transition conditions which are elements of $\mathcal{K}$, and $I(S) \subseteq V(S)$ are the initial states.[7] Also, $E(S) = \{e \in V(S) \times V(S) \mid M_\mathcal{K}(e) \neq 0\}$ is the set of directed edges connecting pairs of vertices of $S$. $M_\mathcal{K}(e)$, which is an abbreviation for $M_\mathcal{K}(S)(e)$, is the transition condition of $M_{\mathcal{K}}(S)$ corresponding to edge $e$. Note that we omit edges labeled "0." By our definition, an edge whose transition condition is 0 does not exist. We can alternatively denote $M_{\mathcal{K}}(e)$ as $M_{\mathcal{K}}(v_i, v_j)$ for the transition condition corresponding to the edge going from vertex $v_i$ to vertex $v_j$. For example, in Figure 3, $M_{\mathcal{K}}((\text{TRANSMITTING, PAUSING}))$ is (I-receive $\wedge$ L-pause).

---

7. There should also be an *output subalgebra*, as in Kurshan (1994). This would help distinguish an agent's own actions from those of other agents. However it is omitted here for notational simplicity.

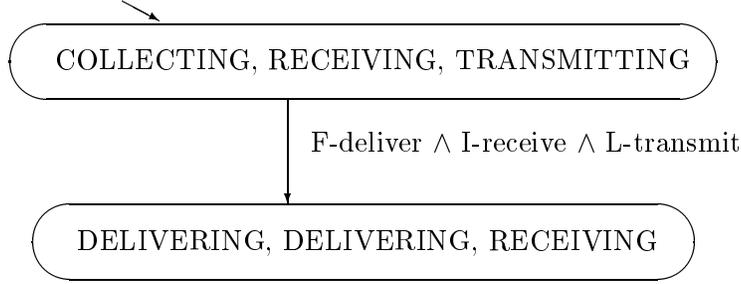

Figure 4: Part of the product plan for agents F, I, and L.

Figures 2 and 3 illustrate these FSA definitions. There are FSA plans for three agents, F, I, and L with vertices, edges, and transition conditions. An incoming arrow to a state, not from any other state, signifies that this is an initial state.

A multiagent plan is formed from single agent plans by taking the *tensor product* (also called the "synchronous product" or simply "product") of the FSAs corresponding to the individual plans. Formally, the tensor product is defined as:

$$\otimes_{i=1}^{n} S_i = (\times V(S_i), \otimes_i M(S_i), \times I(S_i))$$

where $\times$ is the Cartesian product, and the tensor product $M(S_1) \otimes ... \otimes M(S_n)$ of $n$ transition matrices is defined as $M(S_1) \otimes ... \otimes M(S_n)((v_1, v_1'), ..., (v_n, v_n')) = M(S_1)(v_1, v_1') \wedge ... \wedge M(S_n)(v_n, v_n')$ for $(v_1, v_1') \in E(S_1), ..., (v_n, v_n') \in E(S_n)$. In words, the product FSA is formed by taking the Cartesian product of the vertices and the intersection of the transition conditions. Initial states of the product FSA are tuples formed from the initial states of the individual FSAs.

The product FSA models a set of synchronous FSAs. The Boolean algebra corresponding to the product FSA is the product algebra. For Figures 2 and 3, to formulate the FSA S modeling the entire multiagent plan, we take the tensor product S = F $\otimes$ I $\otimes$ L of the three FSAs. For this tensor product, $I(S) = \{(\text{COLLECTING, RECEIVING, TRANSMITTING}), (\text{COLLECTING, RECEIVING, PAUSING}), (\text{COLLECTING, RECEIVING, RECEIVING})\}$. Part of the tensor product FSA is shown in Figure 4.

Next we define the *language* of an FSA, which is the set of all action sequences permitted by the FSA plan. To do this, we first define a *string*, which is a sequence of actions (atoms). Formally, a string $\mathbf{x}$ is an infinite-dimensional vector, $(x_0, ...) \in \Gamma(\mathcal{K})^{\omega}$, i.e., a string is an infinite ($\omega$) length sequence of actions (where $\mathcal{K}$ is the Boolean algebra used by $S$). A *run $\mathbf{v}$ of string $\mathbf{x}$* is a sequence $(v_0, ...)$ of vertices such that $\forall i, \ x_i \wedge M_{\mathcal{K}}(v_i, v_{i+1}) \neq 0$, i.e., $x_i \preceq M_{\mathcal{K}}(v_i, v_{i+1})$ because the $x_i$ are atoms. In other words, a run of a string is the sequence of vertices visited in an FSA when the string satisfies the transition conditions along the edges.

The language of FSA $S$ is defined as:

$$\mathcal{L}(S) \ = \ \{\mathbf{x} \ \in \ \Gamma(\mathcal{K})^{\omega} \mid \mathbf{x} \text{ has a run } \mathbf{v} = (v_0, ...) \text{ in } S \text{ with } v_0 \in I(S) \ \}$$

Such a run is called an *accepting run*, and $S$ is said to *accept* string $\mathbf{x}$. Any requirement on accepting runs of an FSA are what is called the FSA *acceptance criterion*. In this case, the acceptance criterion consists of one condition: accepting runs must begin in an initial state. The verification literature calls these FSAs, which accept infinite-length strings, $\omega$-*automata* (Kurshan, 1994).

A few more definitions are needed. An FSA is *complete* if, for each state $v \in V(S)$, $\sum_{w \in V(S)} M_{\mathcal{K}}(v, w) = 1$. In other words, an FSA is complete if it specifies what state-to-state transition the agent should make for all possible actions taken by the other agents. This is a very reasonable assumption to make because otherwise the agent would not know what to do in some circumstances. An FSA is *deterministic* at state $v$ if $w \neq w' \Rightarrow M_{\mathcal{K}}(v, w) \wedge M_{\mathcal{K}}(v, w') = 0$. In other words, the choice of action uniquely determines which edge will be taken from a state. An FSA is deterministic if it is deterministic at each of its states. Unless otherwise stated, it is assumed here that all FSAs are complete and deterministic. The restriction to deterministic FSAs is not a major problem because for every nondeterministic FSA there is a deterministic one accepting the same language (Kurshan, 1994).

We also need the definition of a *cycle* in a graph. Model checking typically consists of looking for cycles, as described in Section 3.3. A *path* in FSA $S$ is a sequence of vertices $\mathbf{v} = (v_0, ..., v_n) \in V(S)^{n+1}$, for $n \geq 1$ such that $(v_i, v_{i+1}) \in E(S)$ for $i = 0, ..., n-1$, i.e., $M_{\mathcal{K}}(v_i, v_{i+1}) \neq 0$. If $v_n = v_0$, then $\mathbf{v}$ is a cycle. Each cycle in an FSA plan allows the possibility that the agent can infinitely often, or as long as desired, revisit the vertices of the cycle. It also implies that a substring can be repeated indefinitely.

We next illustrate some of these definitions. An example string in the language of FSA S, the multiagent FSA that is the product of F, I, and L, is

((F-collect $\wedge$ I-receive $\wedge$ L-transmit),

(F-deliver $\wedge$ I-receive $\wedge$ L-receive),

(F-deliver $\wedge$ I-receive $\wedge$ L-transmit),

(F-deliver $\wedge$ I-deliver $\wedge$ L-receive), ...).

This is a sequence of atoms of S. A run of this string is

((COLLECTING, RECEIVING, TRANSMITTING),

(DELIVERING, RECEIVING, RECEIVING),

(DELIVERING, RECEIVING, TRANSMITTING),

(DELIVERING, DELIVERING, RECEIVING),

(COLLECTING, RECEIVING, RECEIVING), ...).

All FSAs in Figures 2 and 3 are complete and deterministic. For example, in Figure 2, rover I can only take action I-deliver from its DELIVERING state. However every possible action choice of L determines a unique next state for I from DELIVERING. For example, if L takes L-transmit then I must stay in state DELIVERING, and if L takes L-receive or L-pause then I must go to state RECEIVING.

## 3.2 Properties

Now that we have presented the FSA formalism used for agent plans, we can address the question of how to formalize properties. For verification, properties are typically expressed either as FSAs (for automata-theoretic verification) or in temporal logic. Here, we assume *linear* temporal logic. In other words, we assume that time proceeds linearly and we do not consider simultaneous possible futures. Using the algorithm of Vardi and Wolper (1986), one can convert any linear temporal logic formula into an automaton (because automata are more expressive than linear temporal logic). Both representations are used here. To simplify our proofs in Section 4, properties are expressed in temporal logic. For some of the incremental reverification methods in Section 6, we use automata-theoretic methods with an FSA representation for the property.

Let us begin by defining temporal logic properties. Many of the definitions are based on Manna and Pnueli (1991). To bridge the gap between automata (for plans) and temporal logic (for properties), we need to define a *computational state* (*c-state*). A computation is an infinite sequence of temporally-ordered atoms, i.e., a string. A c-state is an atom in a computation. In other words, it is a (single or multiagent) action that occurs at a single time step in a computation. We continue to refer to an automaton state as simply a "state."

$P$ is a property that is true (false) for an FSA $S$. $S \models P$ ($S \not\models P$), if and only if $P$ is true for every string in the language $\mathcal{L}(S)$ (false for some string in $\mathcal{L}(S)$). The notation $\mathbf{x} \models P$ ($\mathbf{x} \not\models P$) means string $\mathbf{x}$ satisfies (does not satisfy) property $P$, i.e., the property holds (does not hold) for $\mathbf{x}$. Before defining what it means for properties to be true (i.e., hold) for a string, we first define what it means for a formula that is a Boolean expression to be true at a c-state. A *c-state formula* $p$ is true (false) at c-state $x_i$, i.e., $x_i \models p$ ($x_i \not\models p$) if and only if $x_i \preceq p$ ($x_i \not\preceq p$), i.e., $x_i \wedge p \neq 0$ ($= 0$) because $p$ is a Boolean expression with no variables on the same Boolean algebra used by FSA $S$, and $x_i$ is an atom of that algebra. For example, F-collect $\models$ (F-collect $\vee$ F-deliver) for c-state F-collect and c-state formula (F-collect $\vee$ F-deliver). One can also talk about a c-state formula being true or false for an atom, since a c-state is an atom.

A c-state formula $p$ is true or false in particular c-states of a string. Property $P$ is defined in terms of $p$, and is true or false of an entire string. In particular, $\mathbf{x} \models P$ or $\mathbf{x} \not\models P$ for the string $\mathbf{x}$.

We focus on two property classes that are among those most frequently encountered in the verification literature: Invariance and Response properties. Invariance and Response properties are likely to be useful for agents. For the case of a single agent ($SIT_{1agent}$), Invariance properties can express the requirement that a particular action never be executed.[8] Response properties are also useful for a single agent. They can be used to verify that a pair of the agent's actions will occur in the correct order (i.e., a "response" always follows a "trigger") in the plan. In the context of multiple agents ($SIT_{1plan}$ or $SIT_{multplans}$) Invariance properties express the need for parallel multiagent coordination. In particular, they express that multiple agents should not simultaneously perform some conflicting set of actions. Response properties express the need for sequential multiagent coordination. For example, they can express the requirement that one agent's action must follow in response to a particular "triggering" action of another agent.

---

8. This could alternatively be implemented as a run-time check, but then there would be no assurance that the plan without the action is a good one, for example, in terms of how well the revised plan satisfies the agent's goals (perhaps captured in a "fitness function"). Alternatively, the action (atom) could be omitted from the set of actions $\Gamma(\mathcal{K})$. But in general one may not wish to rule out actions, in case the situation and/or properties might change.

Here, we only present informal definitions of these properties; the formal definitions are in Appendix B. An Invariance property $P = \Box \neg p$ ("Invariant not $p$") is true of a string if $p$ is "never" true, i.e., if $p$ is not true in any c-state of the string. $P = \Box(p \rightarrow \Diamond q)$ is a Response property, where $\Diamond$ means "eventually." We call $p$ the "trigger" and $q$ the "response." A Response formula states that every trigger is eventually (in finite time) followed by a response.

To illustrate these property types, we continue the rovers and lander example. The property P1 from Section 2, which states that it should always be the case that I does not deliver at the same time that L is transmitting, is formally expressed as an Invariance property P1 defined as: P1 = $\Box$ ($\neg$(I-deliver $\land$ L-transmit)). Property P2 from Section 2, which states that if F-deliver has occurred then eventually L will execute L-receive, is an example of a Response property. This is expressed in temporal logic as P2 = $\Box$ (F-deliver $\rightarrow \Diamond$ L-receive).

Next consider the FSA representation for properties. As will be explained in Section 3.3 on verification, what we really need to express for automata-theoretic verification is the negation of the property, i.e., $\neg P$. Strings in the language of FSA $\neg P$ violate property $P$. In this paper, we assume that $\neg P$ is expressed using the popular Büchi $\omega$-automaton (Büchi, 1962). We decided to use the Büchi FSA because one of the simplest and most elegant model checking algorithms in the literature assumes this type of FSA for the property, and we use that algorithm (see Subsections 3.3 and 6.1). A Büchi automaton is defined to be a four-tuple $S = (V(S), M_{\mathcal{K}}(S), I(S), B(S))$, where $B(S) \subseteq V(S)$ is a set of "bad" states. To define the language of a Büchi automaton, we require the following preliminary definition. For a run $\mathbf{v}$ of FSA $S$, $\mu(\mathbf{v}) = \{v \in V(S) \mid v_i = v \text{ for infinitely many } v_i\text{s in run } \mathbf{v}\}$. In other words, $\mu(\mathbf{v})$ equals the set of all vertices of $S$ that occur infinitely often in the run $\mathbf{v}$. Then for a Büchi automaton $S$, $\mathcal{L}(S) = \{\mathbf{x} \in \Gamma(K)^\omega \mid \mathbf{x} \text{ has a run } \mathbf{v} = (v_0, ...) \text{ in } S \text{ with } v_0 \in I(S) \text{ and } \mu(\mathbf{v}) \cap B(S) \neq \emptyset\}$. In other words, the Büchi automaton has an acceptance criterion that requires visiting some bad state infinitely often, as well as beginning in an initial state.

An example deterministic Büchi FSA for $\neg$P1, where Invariance property P1 = $\Box \neg$(I-deliver $\land$ L-transmit), is in Figure 5 (on the left) with $B(\neg$P1$) = \{2\}$. Note that visiting a state in $B(\neg$P1$)$ infinitely often implies Büchi acceptance, and because the FSA expresses the negation of the property, visiting a "bad" state in $B(\neg$P1$)$ infinitely often is undesirable. From Figure 5 we can see that any string that includes (I-deliver $\land$ L-transmit) will visit state 2 infinitely often, and $B(\neg$P1$) = \{2\}$. Thus any string that starts in state 1 and includes (I-deliver $\land$ L-transmit) is in $\mathcal{L}(\neg$P1$)$ and therefore violates property P1.

Next consider Response properties of the form $\Box(p \rightarrow \Diamond q)$. For this paper, the only type of FSA that we need for verifying Response properties is the very simple deterministic Büchi FSA for the negation of a "First-Response" property.[9] (Determinism is needed for our efficient internal representation. See Subsection 6.1.) A First-Response property checks whether the *first* trigger $p$ in every string is followed by a response $q$. Figure 5 (on the right) shows a Büchi FSA for the First-Response property corresponding to ¬P2, where property P2 = □ (F-deliver → ◇ L-receive). For this FSA, $B(¬P2) = \{2\}$. Any string whose accepting run visits state 2 infinitely often will include the first trigger and not the response that should follow it. As discussed in Subsection 6.5, verifying First-Response properties can in some circumstances (including all of our experiments) be equivalent to verifying the full Response property □$(p → ◇q)$. Henceforth, when we use the term "Response" this is assumed to include both the full Response and the First-Response versions.

---

9. A straightforward inductive argument shows that it is not possible to construct a deterministic Büchi automaton with a finite number of states for the negation of the full Response property $\Box(p \rightarrow \Diamond q)$ (Mahesh Viswanathan, personal communication).

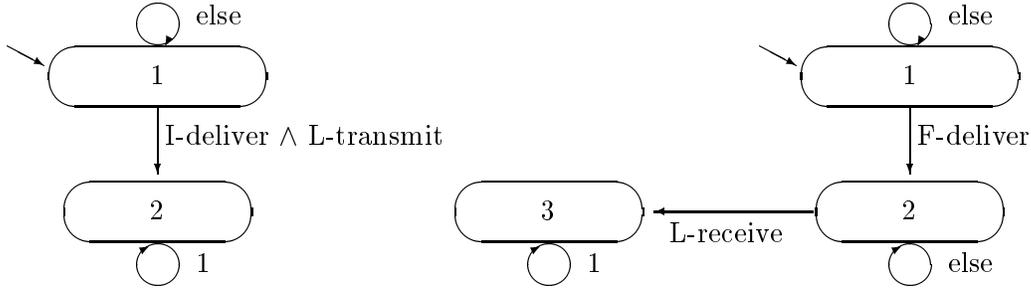

Figure 5: Invariance property ¬P1 (left) and the First-Response version of property ¬P2 (right) as Büchi FSAs, where $B(S) = \{2\}$ for both automata.

## 3.3 Model Checking for Verification

Now that we have our representations for plans and properties, it is possible to describe model checking, i.e., for plan $S$ and property $P$ determining whether $S \models P$. First, however, we need to begin with two essential definitions of accessibility: accessibility of one vertex from another, and accessibility of an atom from a vertex.

**Definition 1** Vertex $v_n$ is *accessible from* vertex $v_0$ if and only if there exists a path from $v_0$ to $v_n$.

**Definition 2** Atom $a_{n-1} \in \Gamma(\mathcal{K})$ is *accessible from* vertex $v_0$ if and only if there exists a path from $v_0$ to $v_n$ and $a_{n-1} \preceq M_{\mathcal{K}}(v_{n-1}, v_n)$.

Accessibility from initial states is central to model checking. The reason is the following. Recall from Section 3.2 that property $P$ is true (false) for an FSA $S$, (i.e., $S \models P$ ($S \not\models P$)), if and only if $P$ is true for every string in the language $\mathcal{L}(S)$ (false for some string in $\mathcal{L}(S)$). By definition, every string in the language has an accepting run. Therefore, it is only necessary to verify the property for strings that have an accepting run. By definition, every accepting run begins with an initial state. Therefore, every state in an accepting run is accessible from an initial state, and every atom (c-state) in a string of the language is accessible from an initial state. Clearly, the only states and atoms that need to be involved in verification are those accessible from initial states.

Invariance properties can be re-expressed in terms of accessibility. Invariance property □¬$p$ could be restated as saying that there does not exist any atom $a$, where $a \preceq p$, that is accessible from an initial state. It is much more difficult to express Response properties succinctly using accessibility. Nevertheless, accessibility plays a key role in verifying all properties, as will be seen shortly.

There are a number of ways to perform model checking, but here we focus on two. The first method is specifically tailored for one class of properties; the second is sufficiently general for use in verifying many classes of properties. The rationale for choosing a specific and a general algorithm is that this allows for a comparison to determine the computational efficiency gained by property-specific tailoring (see Subsection 6.5). In this section, we give high-level sketches of these two model checking algorithms. The full algorithms are in Section 6.

The first algorithm is a very simple and efficient method tailored for Invariance properties $P = \Box \neg p$. For every initial state $v_i$, this method begins at $v_i$ and visits every atom $a_j$ accessible from $v_i$. If this atom has not already been checked, it checks to see whether $a_j \preceq p$. If $a_j \preceq p$, then this is considered a verification failure. If there are no failures, verification succeeds.

The second method, automata-theoretic (AT) model checking, is very popular in the verification literature (e.g, see Vardi and Wolper, 1986) and can be used to verify *any* property expressible as a finite-state automaton. It is used here for First-Response properties. In AT model checking, asking whether $S \models P$ is equivalent to asking whether $\mathcal{L}(S) \subseteq \mathcal{L}(P)$ for property $P$. This is equivalent to $\mathcal{L}(S) \cap \overline{\mathcal{L}(P)} = \emptyset$ (where $\overline{\mathcal{L}(P)}$ denotes the complement of $\mathcal{L}(P)$), which is algorithmically tested by first taking the tensor product of the plan FSA $S$ and the FSA corresponding to $\neg P$ (i.e., $S \otimes \neg P$). The FSA corresponding to $\neg P$ accepts $\overline{\mathcal{L}(P)}$. The tensor product implements language intersection. The algorithm then determines whether $\mathcal{L}(S \otimes \neg P) \neq \emptyset$, which implies $\mathcal{L}(S) \cap \overline{\mathcal{L}(P)} \neq \emptyset$ ($S \not\models P$). This determination is implemented as a check for cycles in the product FSA $S \otimes \neg P$ that are accessible from some initial state and that satisfy any other conditions in the FSA acceptance criterion. Recall that a cycle is a sequence of vertices $(v_0, ..., v_n)$ such that $v_n = v_0$. A cycle is accessible from an initial state if one of its vertices is accessible from the initial state. A cycle that is accessible from an initial state and that satisfies the FSA acceptance criterion implies a nonempty language. This is because a string is in the language of an FSA if it is an infinite-length sequence of actions satisfying the FSA acceptance criterion, which always includes the requirement that its accepting run must begin in an initial state. All infinite behavior eventually ends up in a cycle because the FSA has a finite number of states.

Therefore, to be certain that the language is nonempty, it is necessary to determine whether any accessible cycle satisfies the FSA acceptance criterion. The criterion of interest is the Büchi criterion, for the following reason. It is assumed here that the negation of the property ($\neg P$) is expressed as a Büchi automaton. This implies that the FSA being searched, i.e., $S \otimes \neg P$, is also a Büchi automaton, because taking the tensor product preserves this criterion. The final check of this algorithm is whether an accessible cycle in $S \otimes \neg P$ satisfies the Büchi acceptance criterion, because in that case the language is not empty. A product state $s$ is in $B(S \otimes \neg P)$ whenever it has a component state in $B(\neg P)$, e.g., (COLLECTING, RECEIVING, RECEIVING, 2) is in $B(S \otimes \neg P2)$ for property P2 because its fourth component is state 2 of $B(\neg P2)$. According to the Büchi acceptance criterion, visiting a state $v \in B(S \otimes \neg P)$ infinitely often (assuming $v$ is accessible from an initial state) implies $\mathcal{L}(S \otimes \neg P) \neq \emptyset$. This will happen if $v$ is part of an accessible cycle. In that case, $S \not\models P$ and verification fails. Otherwise, if no accessible product state $v \in B(S \otimes \neg P)$ is visited infinitely often (i.e., it is not in a cycle), then $\mathcal{L}(S \otimes \neg P) = \emptyset$ and therefore $\mathcal{L}(S) \subseteq \mathcal{L}(P)$, i.e., $S \models P$ and verification succeeds. A relatively efficient algorithm for AT verification from the literature is presented in Section 6.

## 3.4 Machine Learning to Adapt Plans

Given plan $S$ and property $P$, model checking determines whether $S \models P$. Next we consider the case of learning, which is a change to $S$. This subsection addresses the issue of how a learning operator can affect a plan $S$ to generate a new plan $S'$.

We begin by presenting a taxonomy of FSA learning operators. It is likely that any learning method for complete deterministic FSAs will be composed of one or more of these operators. Nothing about our approach requires evolutionary learning per se; however to make the discussion concrete, this is the form of learning that is assumed here. In the context of evolutionary algorithms, the FSA learning operators are perturbations, such as mutations, applied to the FSAs.

```
Procedure EA
t = 0; /* initial generation */
initialize_population(t);
evaluate_fitness(t);
until termination-criterion do
        t = t + 1; /* next generation */
        select_parents(t);
        perturb(t);
        evaluate_fitness(t);
enduntil
end procedure
```

Figure 6: The outline of an evolutionary algorithm.

We assume that learning occurs in two phases: the offline and online phases (see Figure 1). During the offline phase, each agent starts with a randomly initialized population of candidate FSA plans. This population is evolved using the evolutionary algorithm outlined in Figure 6. The main loop of this algorithm consists of selecting parent plans from the population, applying perturbation operators to the parents to produce offspring, evaluating the fitness of the offspring, and then returning the offspring to the population if they are sufficiently fit. After this evolution, verification and repair are done to these initially generated plans.

At the start of the online phase, each agent selects one "best" (according to its "fitness function") plan from its population for execution. The agents are then fielded and plan execution is interleaved with learning (adaptation), reverification, and plan repair as needed. The purpose of learning during the online phase is to fine-tune the plan and adapt it to keep pace with a gradually shifting environment, since normally real-world environments are not static. The evolutionary algorithm of Figure 6 is also used during this phase, but the assumption is a population size of one and incremental learning (i.e., one learning operator applied per FSA per generation). This is practical for situations in which the environment changes gradually, rather than radically.

Formally, a machine learning operator $o : S \rightarrow S'$ changes a (product or individual) FSA $S$ to post-learning FSA $S'$. A mapping between two automata $S$ and $S'$ is defined as a mapping between their elements (Bavel, 1983). At the highest level, we can subdivide the learning operators according to the elements of the FSA that they alter:

- One class of operators adds, deletes, or moves edge transition conditions. In other words, $o : M_{\mathcal{K}}(S) \rightarrow M_{\mathcal{K}}(S')$.

- Another class of operators adds, deletes, or moves edges, i.e., $o : E(S) \rightarrow E(S')$.

- The third class of operators adds or deletes vertices, along with their edges, i.e., $o : V(S) \rightarrow V(S')$ and $o : E(S) \rightarrow E(S')$.

- The fourth class of operators changes the Boolean algebra used in the transition conditions, i.e., $o : \mathcal{K} \rightarrow \mathcal{K}'$.

Here, we do not define operators that add or delete states. In other words, we do not address the third class of operators. The reason is that with the type of FSAs used here, adding or deleting a state does not, in itself, affect properties. It is what we do with the edges to/from a state and their transition conditions that can alter whether a property is true or false for a plan. This is because properties are true or false for comp-states (atoms) rather than for FSA states. Furthermore, this paper does not address changes to the Boolean algebra, which is the fourth class of operators. This class of operators, which includes abstractions, is addressed in Gordon (1998).

Therefore we are focusing on the first and second classes of operators. We define operator schemas, rather than operators. A machine learning operator schema applies to unspecified (variable) vertices, edges, and transition conditions. When instantiated with particular vertices, edges, and transition conditions, it becomes a machine learning operator. In order to avoid tedium, the operator schema definitions consider only the relevant parts of the FSA, e.g., those parts that get altered. There is an implicit assumption that all unspecified parts of the FSA remain the same after operator application. There is also an assumption that the learner ensures that all operators keep the automaton complete and deterministic.

The operators can be seen in the taxonomy (partition) of Figure 7. We define each of the corresponding operator schemas as follows, beginning with the most general one, called $o_{change}$, which changes edge transition conditions:

**Operator Schema 1 ($o_{change}$)** *Let $S$ be an FSA with Boolean algebra $\mathcal{K}$, and let $o_{change} : S \rightarrow S'$. Then we define $o_{change} : M_{\mathcal{K}}(S) \rightarrow M_{\mathcal{K}}(S')$. In particular, suppose $z \preceq M_{\mathcal{K}}(v_1, v_2)$, $z \neq 0$, for $(v_1, v_2) \in E(S)$ and $z \npreceq M_{\mathcal{K}}(v_1, v_3)$ for $(v_1, v_3) \in E(S)$. Then $o_{change}(M_{\mathcal{K}}(v_1, v_2)) = M_{\mathcal{K}}(v_1, v_2) \wedge \neg z$ (step 1) and/or $o_{change}(M_{\mathcal{K}}(v_1, v_3)) = M_{\mathcal{K}}(v_1, v_3) \vee z$ (step 2). In other words, $o_{change}$ may consist of two steps: the first to remove condition $z$ from edge $(v_1, v_2)$, and the second to add (the same) condition $z$ to edge $(v_1, v_3)$. Alternatively, $o_{change}$ may consist of only one of these two steps.*

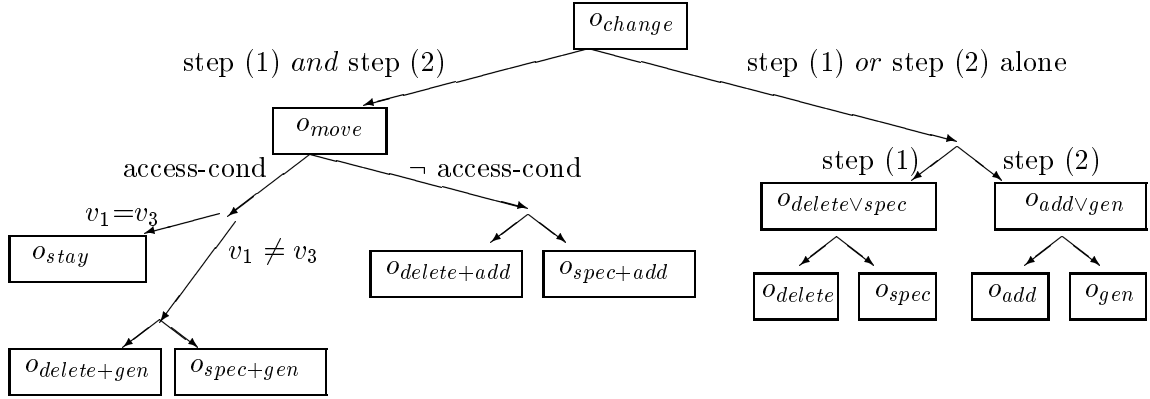

Figure 7: Taxonomy (partition) of learning operators.

All of the remaining operators are easier to describe in terms of a set of four primitive operators. Therefore, we next define these four primitives, which are one-step operators that are special cases of $o_{change}$ and appear at the bottom right as leaves in the hierarchy of Figure 7. The first two primitive operators delete ($o_{delete}$) and add ($o_{add}$) edges. We define $o_{delete}$ to delete edge $(v_1, v_2)$ with the operator schema:

**Operator Schema 2 ($o_{delete}$)** *Let $S$ be an FSA with Boolean algebra $\mathcal{K}$, and let $o_{delete}$ : $S \to S'$ be defined with $o_{delete} : E(S) \to E(S) \setminus \{(v_1, v_2)\}$ for deleted edge $(v_1, v_2)$ of $S$. Recall that a nonexistent edge has transition condition 0. Operator $o_{delete}$ could therefore be considered a special case of $o_{change}$ that consists only of step (1) and an additional condition that must be met, namely, that $o_{delete}(M_{\mathcal{K}}(v_1, v_2)) = (M_{\mathcal{K}}(v_1, v_2) \wedge \neg z) = 0$.*

We define $o_{add}$ to add edge $(v_1, v_3)$ with the operator schema:

**Operator Schema 3 ($o_{add}$)** *Let $S$ be an FSA with Boolean algebra $\mathcal{K}$, and let $o_{add} : S \to S'$ be defined with $o_{add} : E(S) \to E(S) \cup \{(v_1, v_3)\}$ for added edge $(v_1, v_3)$ of $S$. Operator $o_{add}$ could be considered a special case of $o_{change}$ that consists only of step (2) and the additional condition that $M_{\mathcal{K}}(v_1, v_3) = 0$ prior to applying $o_{add}$.*

The other two primitive operators are specialization ($o_{spec}$) and generalization ($o_{gen}$). Specialization and generalization are operators commonly found in the machine learning literature, e.g., see Michalski (1983). In the context of an FSA, specialization lowers the level of a particular state-to-state transition condition in the partial order $\preceq$, whereas generalization raises it, as in Mitchell's Version Spaces (Mitchell, 1978). In particular, a transition condition can be specialized with a meet and can be generalized with a join, which is analogous to adding a conjunct to specialize and a disjunct to generalize as in Michalski (1983).

Formally, we define specialization and generalization, respectively, as follows:

**Operator Schema 4 ($o_{spec}$)** *Let $S$ be an FSA with Boolean algebra $\mathcal{K}$, and let $o_{spec} : S \to S'$. Then we can define $o_{spec} : M_{\mathcal{K}}(S) \to M_{\mathcal{K}}(S')$, where $o_{spec}(M_{\mathcal{K}}(v_1, v_2)) = M_{\mathcal{K}}(v_1, v_2) \wedge$* $\neg z$, for some $z \in \mathcal{K}$, $z \neq 0$. Operator $o_{spec}$ could be considered a special case of $o_{change}$ that consists only of step (1) and the additional two conditions $o_{spec}(M_{\mathcal{K}}(v_1, v_2)) = (M_{\mathcal{K}}(v_1, v_2) \wedge \neg z) \neq 0$ (i.e., $o_{spec} \neq o_{delete}$), and $M_{\mathcal{K}}(v_1, v_2) \neq \neg z$ (since otherwise $o_{spec}$ has no effect).

**Operator Schema 5 ($o_{\mathbf{gen}}$)** *Let $S$ be an FSA with Boolean algebra $\mathcal{K}$, and let $o_{gen} : S \to S'$. Then we can define $o_{gen} : M_{\mathcal{K}}(S) \to M_{\mathcal{K}}(S')$, where $o_{gen}(M_{\mathcal{K}}(v_1, v_3)) = M_{\mathcal{K}}(v_1, v_3) \vee z$, for some $z \in \mathcal{K}$, $z \neq 0$. Operator $o_{gen}$ could be considered a special case of $o_{change}$ that consists only of step (2) and the two additional conditions that $M_{\mathcal{K}}(v_1, v_3) \neq 0$ (i.e., $o_{gen} \neq o_{add}$) and $(M_{\mathcal{K}}(v_1, v_3) \wedge z) = 0$ (because otherwise $z$ adds redundancy) prior to $o_{gen}$.*

Next, 10 learning operators are defined from these four primitives. Below $o_{change}$ in the operator hierarchy of Figure 7 are two subtrees. The right subtree consists of one-step operators, and the left subtree consists of two-step operators. We define the two one-step operators just below $o_{change}$ first (since we just defined the primitive operators below them):

**Operator Schema 6 ($o_{\mathbf{delete} \vee \mathbf{spec}}$)** *This operator consists of applying either of the primitive operators $o_{delete}$ or $o_{spec}$.*

**Operator Schema 7 ($o_{\mathbf{add} \vee \mathbf{gen}}$)** *This operator consists of applying either of the primitive operators $o_{add}$ or $o_{gen}$.*

It is relevant at this point to introduce two more operators that are not in the hierarchy of Figure 7. They are not in the hierarchy because they are merely minor variants of $o_{delete \vee spec}$ and $o_{add \vee gen}$ and they do not belong strictly below our most general operator $o_{change}$. These operators are introduced here because they are very useful and also because they are guaranteed to preserve completeness of FSAs. In other words, if the FSA is complete prior to applying these operators then it will be complete after applying them. Recall from Section 2 that each FSA state is associated with a set of allowable actions that may be taken from that state. These operators delete or add an action from the set of allowable actions from a state:

**Operator Schema 8 ($o_{\mathbf{delete-action}}$)** *Delete an allowable action from a state $v_1$ by one or more applications of operator $o_{delete \vee spec}$. Each application may be to a different outgoing edge from $v_1$.*

**Operator Schema 9 ($o_{\mathbf{add-action}}$)** *Add an allowable action from a state $v_1$ by one or more applications of operator $o_{add \vee gen}$. Each application may be to a different outgoing edge from $v_1$.*

To understand why $o_{delete-action}$ consists of one or more applications of $o_{delete \vee spec}$, consider the following example. In Figure 2, deleting F-collect as an allowable action from F's COLLECTING state results in F-deliver being the only allowable action from that state. Furthermore, this results in the edge (COLLECTING, COLLECTING) being deleted and the edge (COLLECTING, DELIVERING) being specialized. The reasoning is similar for why $o_{add-action}$ is one or more applications of $o_{add \vee gen}$.

The remaining operators, which are all of the operators on the left subtree of $o_{change}$ in Figure 7, consist of two steps: the first to remove condition $z$ from edge $(v_1, v_2)$, and the second to add (the same) condition $z$ to edge $(v_1, v_3)$. The first step consists of applying one primitive operator, and the second step consists of applying another primitive operator. Every one of the following operators preserves determinism and completeness of the FSAs. In other words, if the FSA is deterministic and complete prior to operator application then it will be deterministic and complete afterwards.

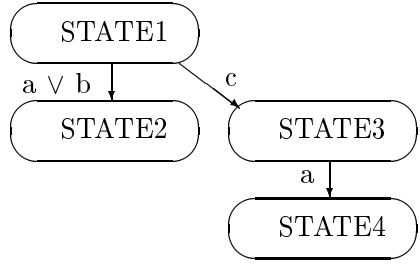

Figure 8: Moving transition conditions between edges.

**Operator Schema 10 ($o_{\mathbf{move}}$)** *This operator schema is identical to that of $o_{change}$, with one exception. Replace "and/or" with "and" in the definition. In other words, we have $o_{move}(M_\mathcal{K}(v_1, v_2)) = M_\mathcal{K}(v_1, v_2) \wedge \neg z$ and $o_{move}(M_\mathcal{K}(v_1, v_3)) = M_\mathcal{K}(v_1, v_3) \vee z$ for some $(v_1, v_2), (v_1, v_3) \in E(S)$. Therefore $o_{move}$ moves $z$ from one edge to another.*

All of the remaining operators are special cases of $o_{move}$. We begin with the right subtree of $o_{move}$:

**Operator Schema 11 ($o_{\mathbf{delete+add}}$)** *Apply $o_{delete}$ to edge $(v_1, v_2)$ and then apply $o_{add}$ to edge $(v_1, v_3)$.*

An example of $o_{delete+add}$, using Figure 8, is to delete edge (STATE1, STATE3) (i.e., make $M_\mathcal{K}$ (STATE1, STATE3) $= 0$) and add a new edge (STATE1, STATE1) with transition condition $M_\mathcal{K}$ (STATE1, STATE1) $=$ c.

**Operator Schema 12 ($o_{\mathbf{spec+add}}$)** *Apply $o_{spec}$ to edge $(v_1, v_2)$ and then apply $o_{add}$ to edge $(v_1, v_3)$.*

For example, using Figure 8, we can move "b" from edge (STATE1, STATE2) to a newly created edge (STATE1, STATE1) to make $M_\mathcal{K}$ (STATE1, STATE2) $=$ a $\wedge \neg$b and $M_\mathcal{K}$ (STATE1, STATE1) $=$ b. This is specialization of the condition on edge (STATE1, STATE2) followed by addition of edge (STATE1, STATE1).

Next consider the left subtree of $o_{move}$. At this point, it is relevant to examine the reason for the split into the two subtrees of $o_{move}$. All of the operators in the left subtree satisfy a condition that is called the "accessibility condition." This condition states that prior to learning (and also after learning), if vertex $v_1$ is accessible from some initial state then vertex $v_3$ is guaranteed to also be accessible from that initial state. The reason for this partition will become clear in Subsection 4.2, where we show that a theorem holds for the two-step operators if and only if the accessibility condition is true. The reason that the two operators in the right subtree of $o_{move}$ fail to satisfy the accessibility condition is due to their having $o_{add}$ as their second step. The definition of $o_{add}$ states that $M_{\mathcal{K}}(v_1, v_3) = 0$ prior to operator application, and therefore we have no guarantee of $v_3$'s accessibility, given that $v_1$ is accessible from an initial state. The following are the definitions of the operators for which the accessibility condition is true:

**Operator Schema 13 ($o_{delete+gen}$)** *Apply $o_{delete}$ to edge $(v_1, v_2)$ and then apply $o_{gen}$ to edge $(v_1, v_3)$.*

As an example, in Figure 8, we can move the condition "a ∨ b" from edge (STATE1, STATE2) to edge (STATE1, STATE3) to make $M_{\mathcal{K}}$ (STATE1, STATE2) = 0 and to make $M_{\mathcal{K}}$ (STATE1, STATE3) = c ∨ a ∨ b. This is deletion of edge (STATE1, STATE2) followed by generalization of the transition condition on edge (STATE1, STATE3).

**Operator Schema 14 ($o_{spec+gen}$)** *Apply $o_{spec}$ to edge $(v_1, v_2)$ and then apply $o_{gen}$ to edge $(v_1, v_3)$.*

As an example, in Figure 8, we can move the disjunct "b" from edge (STATE1, STATE2) to edge (STATE1, STATE3) to make $M_{\mathcal{K}}$ (STATE1, STATE2) = a ∧ ¬b and $M_{\mathcal{K}}$ (STATE1, STATE3) = c ∨ b. This is a specialization of the transition condition on edge (STATE1, STATE2) followed by a generalization of the transition condition on edge (STATE1, STATE3).

**Operator Schema 15 ($o_{stay}$)** *The definition is the same as that of $o_{move}$, with one exception. Replace vertex $v_3$ with vertex $v_1$ everywhere. In other words, the operator consists of moving a condition from edge $(v_1, v_2)$ to edge $(v_1, v_1)$.*

Note that each operator *instantiation* of the schema for $o_{stay}$ will be a special case of one of the following: $o_{delete+add}$, $o_{spec+add}$, $o_{delete+gen}$, or $o_{spec+gen}$. It is considered $o_{stay}$ if and only if on the second step of the operator the transition condition is moved to edge $(v_1, v_1)$. For example, using Figure 8, when we applied operator $o_{spec+add}$ (in the example above) to move the disjunct "b" from edge (STATE1, STATE2) to edge (STATE1, STATE1) to make $M_{\mathcal{K}}$ (STATE1, STATE2) = a ∧ ¬b and $M_{\mathcal{K}}$ (STATE1, STATE1) = b, this could be considered an instantiation of $o_{stay}$, as well as $o_{spec+add}$. Likewise when we applied $o_{delete+add}$ to delete edge (STATE1, STATE3) and add edge (STATE1, STATE1) with "c" as the transition condition, this could also be considered an instantiation of $o_{stay}$.

Operator $o_{stay}$ is an especially useful operator. It makes the reasonable assumption that when an agent no longer wants to transition to another state (e.g., an edge is deleted), the agent just stays in its current state. In other words, the condition for transitioning to another state is transferred to the edge leading back to the current state. For example, suppose rover I becomes stuck at the lander and cannot rendezvous with F for an indeterminate period of time. It could generate a temporary plan (see Figure 2) that keeps I in its DELIVERING state by deleting edge (DELIVERING, RECEIVING) and making $M_{\mathcal{K}}$(DELIVERING, DELIVERING) = 1 (and DELIVERING would have to become an initial state).

Recall that accessibility is a key issue for verification. Now that we have a set of operator schemas, let us consider how these operators affect accessibility from initial states.

Clarifying this will be relevant for understanding both the a priori proofs about property preservation, and the motivation for the incremental reverification algorithms. There are two fundamental ways that our learning operators may affect accessibility: *locally* (abbreviated "L"), i.e., by directly altering the accessibility of atoms or states, or *globally* (abbreviated "G"), i.e., by altering accessibility of states or atoms that could be visited *after* the part of the FSA modified by the learning operator. In particular, any change to the accessibility of $v_1$, $v_2$, $v_3$ or atoms in $M_\mathcal{K}(v_1, v_2)$ or $M_\mathcal{K}(v_1, v_3)$, referenced in the operator definition, is considered local. Changes to accessibility of any other states or atoms is considered global.

As an example of an L (local) change to accessibility, using Figure 8, suppose the agent discovers a new action "d" that it can take. It adds "d" to its action repertoire, as well as to the set of allowable actions from one of the states in its FSA. In particular, the agent decides to allow "d" from STATE1 and decides to apply $o_{gen}$ to the transition condition for (STATE1, STATE3) to get condition "c ∨ d." Then atom "d" was not previously accessible from any initial state, but if we assume STATE1 is accessible from an initial state then the application of $o_{gen}$ made the atom "d" accessible. Using Figure 8 to illustrate a G (global) change to accessibility, suppose we delete edge (STATE1, STATE3) in that figure. Then STATE4, which was previously accessible (because we assume STATE1 is accessible) is no longer accessible. On the other hand, the fact that STATE3 is no longer accessible is a local change.

Now we are ready to summarize what the learning operators can do to accessibility. First, we introduce one more notational convenience. The symbols ↑ and ↓ denote "can increase" and "can decrease," respectively, and ↗̸ and ↙̸ denote "cannot increase" and "cannot decrease," respectively. We use these symbols with G and L, e.g., ↑ G means that a learning operator can (but does not necessarily) increase global accessibility, and ↙̸ L means that an operator cannot decrease local accessibility.

The results for the primitive operators are intuitively obvious:

- $o_{delete}$: ↓ G ↓ L ↗̸ G ↗̸ L

- $o_{spec}$: ↙̸ G ↓ L ↗̸ G ↗̸ L

- $o_{add}$: ↙̸ G ↙̸ L ↑ G ↑ L

- $o_{gen}$: ↙̸ G ↙̸ L ↗̸ G ↑ L

The primitive operators provide answers about changes in accessibility for all of the one-step operators. For the two-step operators (i.e., $o_{move}$ and all operators below it in the hierarchy of Figure 7), we need to consider the *net* effect. For the results in this paper, we only need to focus on one distinction – the difference in the net effect for those operators that satisfy the accessibility condition (i.e., the left subtree of $o_{move}$) versus the net effect for those operators that do not satisfy this condition (i.e., the right subtree). The net effect of those operators that satisfy the accessibility condition is that accessibility (global and local) will never be increased, i.e., ↗̸ G and ↗̸ L. The reason is as follows. By looking at the results for the primitive operators, it is apparent that the first step in these two-step operators can never increase accessibility, because the first step is always $o_{delete}$ or $o_{spec}$. Therefore, to understand the intuition behind this result we need to examine the second step. Consider $o_{delete+gen}$ and $o_{spec+gen}$. Note that $o_{gen}$ does not increase global accessibility ($\not\uparrow$ G), but it can increase local accessibility ($\uparrow$ L). Is $\uparrow$ L a net effect due to the generalization step? Because atoms are being transferred from one outgoing edge of some vertex $v_1$ to another outgoing edge of $v_1$ with these two operators, by definition the local accessibility of those atoms from an initial state will not be increased as a net effect. In other words, the atoms are accessible from an initial state if and only if $v_1$ is, and these two learning operators do not increase the accessibility of $v_1$. Furthermore, by definition $M_\mathcal{K}(v_1, v_3) \neq 0$ prior to learning, so the accessibility of $v_3$ is not increased. We conclude that $\not\uparrow$ L is a net effect.

A similar line of reasoning explains why operator $o_{stay}$ will not increase local accessibility. Operator $o_{stay}$ cannot increase global accessibility, even if it adds an edge, because the only edge that this operator could add is $(v_1, v_1)$. In conclusion, all three operators that satisfy the accessibility condition have a net effect of *not* increasing accessibility ($\not\uparrow$ G and $\not\uparrow$ L). On the other hand, because operators $o_{delete+add}$ and $o_{spec+add}$ have $o_{add}$ as their second step, they can increase accessibility.

Results from lower in the hierarchy of Figure 7 are inherited up the tree. For example, because $o_{delete+add}$ can increase global accessibility, $o_{move}$ can as well. The following is a summary of the relevant results we have so far about how the two-step learning operators can change accessibility. To avoid overwhelming the reader, we present only those results necessary for understanding this paper:

- $o_{stay}$, $o_{delete+gen}$, $o_{spec+gen}$: $\not\uparrow$ G $\not\uparrow$ L

- $o_{delete+add}$, $o_{spec+add}$, $o_{move}$, $o_{change}$: $\uparrow$ G

Before concluding this section, we briefly consider a different partition of the learning operators than that reflected in the taxonomy of Figure 7. This different partition is necessary for understanding the a priori proofs about the preservation of Response properties (in Section 4). For this partition, we wish to distinguish those operators that can introduce at least one new string with an infinitely repeating substring (e.g., (a,b,c,d,e,d,e,d,e,...) where the ellipsis represents infinite repetition of d followed by e) into the FSA language versus those that cannot. Any operator that can add atoms to the transition condition for an edge in a cycle, add an edge to an existing cycle, or add an edge to create a new cycle belongs to the first class (the class that can add such substrings). Thus this first class includes our operators that can create new cycles (e.g., $o_{stay}$ because it can add a new edge $(v_1, v_1)$), as well as our operators that can generalize the transition condition along some edge of a cycle (e.g., $o_{delete+gen}$ because it can generalize $M_\mathcal{K}(v_1, v_1)$). The operators are divided between these two classes as follows:

1. $o_{add}$, $o_{gen}$, $o_{add\lor gen}$, $o_{add-action}$, $o_{stay}$, $o_{delete+gen}$, $o_{spec+gen}$, $o_{delete+add}$, $o_{spec+add}$, $o_{move}$, $o_{change}$

2. $o_{delete}$, $o_{spec}$, $o_{delete\lor spec}$, $o_{delete-action}$

It is important to note that *all* of the two-step operators are in the first class.

At this point we have defined a set of useful operators (via their operator schemas) that one could apply to an FSA plan for adaptation. With these operators, it is possible to improve the effectiveness of a plan, and to adapt it to handle previously unforeseen external and internal conditions. To ensure the usefulness of these learning operators, the learner needs to check that it has not generated a useless plan (i.e., $\mathcal{L}(S) \neq \emptyset$). Although not addressed in this paper, we are currently developing efficient methods for making this check using the knowledge of the learning that was done.

The particular choice of learning operators presented here was motivated by four factors. First, these operators translate into easy-to-implement perturbations of entries in a table, which is the representation of FSAs used in our implementation (see Section 6). Second, these operators were inspired by the literature. For example, generalization and specialization operators are considered fundamental for inductive inference (Michalski, 1983), and deleting/adding FSA edges are effective for evolving FSAs (Fogel, 1996). Third, these operators made practical sense in the context of applications that were considered. Fourth, the particular taxonomies presented here facilitate powerful theoretical and empirical results for reducing the time complexity of reverification, as shown in the remainder of this paper.

## 4. A Priori Results about the Safety of Machine Learning Operators

Subsection 3.4 defined several useful learning operator schemas to modify automaton edges ($o : E(S) \to E(S')$) and the transition conditions along edges ($o : M_{\mathcal{K}}(S) \to M_{\mathcal{K}}(S')$). The results in this section establish which of these operator schemas $o$ are a priori guaranteed to preserve two property classes of interest (Invariance and Response). This section assumes that all learning operators are applied to a single FSA plan, i.e., $SIT_{1agent}$ or $SIT_{1plan}$. Section 5 addresses the translation of the operators applied to a single plan into their effect on a product plan (for $SIT_{multplans}$), and how this affects the results. We begin by formally defining what we mean by "safe machine learning operator."

### 4.1 "Safe" Online Machine Learning

Our objective is to lower the time complexity of reverification. The ideal solution is to identify *safe machine learning* methods (SMLs), which are machine learning operators that are a priori guaranteed to preserve properties (also called "correctness preserving mappings") and require no run-time reverification. For a plan $S$ and property $P$, suppose verification has succeeded prior to learning, i.e., $\forall \mathbf{x}, \quad \mathbf{x} \in \mathcal{L}(S)$ implies $\mathbf{x} \models P$ (i.e., $S \models P$). Then a machine learning operator $o(S)$ is an SML if and only if verification is guaranteed to succeed after learning. In other words, if $S' = o(S)$, then $S \models P$ implies $S' \models P$.

Subsection 4.2 provides results about the a priori safety of machine learning operators. Some of the results in Subsection 4.2 are negative. Nevertheless, although we do not have an a priori guarantee for these learning operators, Section 6 shows that we can perform reverification more efficiently than total reverification from scratch.

### 4.2 Theoretical Results

Let us begin by considering the primitive operators. The results for all primitive operators are corollaries of two fundamental theorems, Theorems 1 and 2, which may not be immediately intuitive. For example, it seems reasonable to suspect that if an edge is deleted somewhere along the path from a trigger to a response, then this could cause failure of a Response property to hold because the response is no longer accessible. In fact, this is not true. What actually happens is that deletions reduce the number of strings in the language. If the original language satisfies the property then so does is the smaller language. Theorem 1 formalizes this.

**Theorem 1** *Let $S'$ be an FSA with Boolean algebra $\mathcal{K}$. Let $S$ be identical to $S'$, but with additional edges, i.e., $o : S' \to S$ is defined as $o : E(S') \to E(S)$, where $E(S') \subseteq E(S)$. Then $\mathcal{L}(S') \subseteq \mathcal{L}(S)$.*

**Proof.** The language may be enlarged by the addition of new edges that have newly learned transition conditions. On the other hand, because every accepting run remains an accepting run regardless of new edges, $\mathbf{x} \in \mathcal{L}(S')$ implies $\mathbf{x} \in \mathcal{L}(S)$, and we are never reducing the size of the language. Therefore, $\mathcal{L}(S') \subseteq \mathcal{L}(S)$. □

The results about the machine learning operator schemas $o_{delete}$ and $o_{add}$ follow as corollaries:

**Corollary 1** *$o_{delete}$ is an SML with respect to any property $P$.*

**Proof.** Assume $S \models P$. Then $\forall \mathbf{x}$, $\mathbf{x} \in \mathcal{L}(S)$ implies $\mathbf{x} \models P$. Define $o_{delete}(S) = S'$. By Theorem 1, $\mathcal{L}(S') \subseteq \mathcal{L}(S)$. Therefore, $\forall \mathbf{x}$, $\mathbf{x} \in \mathcal{L}(S')$ implies $\mathbf{x} \models P$. We conclude that $S' \models P$, i.e., $o_{delete}(S) \models P$. □

To be consistent with Theorem 1, in Corollary 2 only (but not in the rest of the paper), we use $S'$ for the pre-$o_{add}$ FSA and $S$ for the post-$o_{add}$ FSA, i.e., $o_{add}(S') = S$.

**Corollary 2** *$o_{add}$ is not necessarily an SML for any property, including Invariance and Response properties.*

**Proof.** Assume $S' \models P$. Then $\forall \mathbf{x}$, $\mathbf{x} \in \mathcal{L}(S')$ implies $\mathbf{x} \models P$. By Theorem 1, $\mathcal{L}(S') \subseteq \mathcal{L}(S)$. Then we cannot be certain that $S \models P$, i.e., that $o_{add}(S') \models P$. For instance, a counterexample for Invariance property $\square \neg p$ occurs if we add an accessible edge with transition condition $p$. □

Now we consider a priori results for $o_{spec}$ and $o_{gen}$. Again, we begin with a relevant theorem for operator schema $o$.

**Theorem 2** *Let $S'$ be an FSA with Boolean algebra $\mathcal{K}$, and let $o : S' \to S$ be defined as $o : M_{\mathcal{K}}(S') \to M_{\mathcal{K}}(S)$ where $\exists z \in \mathcal{K}$, $z \neq 0$, $(v_1, v_3) \in E(S')$, such that $o(M_{\mathcal{K}}(v_1, v_3)) = M_{\mathcal{K}}(v_1, v_3) \vee z$. Then $\mathcal{L}(S') \subseteq \mathcal{L}(S)$.*

**Proof.** Similar to the proof of Theorem 1. □

**Corollary 3** *$o_{spec}$ is an SML for any property.*

**Proof.** Similar to the proof of Corollary 1 of Theorem 1. □

**Corollary 4** *$o_{gen}$ is not necessarily an SML for any property, including Invariance and Response properties.*

**Proof.** Similar to the proof of Corollary 2 of Theorem 1. □

We can draw the following conclusions from the theorems and corollaries just presented:

- Of the one-step learning operators, those that are guaranteed to be SMLs for any property are $o_{delete}$, $o_{spec}$, and $o_{delete \lor spec}$ (which implies that $o_{delete-action}$ is also an SML for any property).

- We need never be concerned with the first step in a two-step operator. It is guaranteed to be an SML (because $o_{delete}$ or $o_{spec}$ is always the first step).

Next consider theorems that are needed to address the two-step operators. Although we found results for the one-step operators that were general enough to address *any* property, we were unable to do likewise for the two-step operators. Our results for the two-step operators determine whether these operators are necessarily SMLs for Invariance or Response properties in particular. Future work will consider other property classes. The theorems are quite intuitive. The first theorem distinguishes those learning operators that will satisfy Invariance properties from those that will not:

**Theorem 3** *A machine learning operator is guaranteed to be an SML with respect to any Invariance property $P$ if and only if $\not\uparrow$ G and $\not\uparrow$ L are both true (which, for our two-step operators, implies that the operator satisfies the accessibility condition).*

**Proof.** Suppose $\not\uparrow$ G and $\not\uparrow$ L are both true. Let Invariance property $P = \Box \neg p$. Assume $P$ is true of FSA $S$ prior to learning. Then for every string $\mathbf{y} \in \mathcal{L}(S)$, it must be the case that $\neg p$ is true in every c-state of $\mathbf{y}$. If accessibility of atoms is not increased (i.e., $\not\uparrow$ G and $\not\uparrow$ L), then it must be the case that every c-state of every string $\mathbf{x} \in \mathcal{L}(S')$, where $S' = o(S)$, is also a c-state of some string in $\mathcal{L}(S)$. Therefore, for every string $\mathbf{x} \in \mathcal{L}(S')$, it must be the case that $\neg p$ is true in every c-state of $\mathbf{x}$. In other words, moving transition conditions around in an FSA without increasing accessibility will not alter the truth of an Invariance property, which holds in every c-state of every string in the language of the FSA.

Suppose $\uparrow$ G or $\uparrow$ L. Increasing accessibility of atoms implies the possibility of introducing a c-state in some string $\mathbf{x} \in \mathcal{L}(S')$, where $S' = o(S)$, that was not in any string of $\mathcal{L}(S)$. This can cause violation of an Invariance property, as in the counterexample in the proof of Corollary 2. Knowing that $\neg p$ is true in every c-state of every string of $\mathcal{L}(S)$ provides no guarantee that $\neg p$ is true in every c-state of every string of $\mathcal{L}(S')$. □

Since we already have results to cover the one-step operators, we need only consider the two-step operators.

**Corollary 5** *The machine learning operator schemas $o_{delete+gen}$, $o_{spec+gen}$, and $o_{stay}$ are guaranteed to be SMLs with respect to any Invariance property $P$ because for all of these operators $\not\uparrow$ G and $\not\uparrow$ L.*

**Corollary 6** *The machine learning operator schemas $o_{delete+add}$, $o_{spec+add}$, $o_{move}$, and $o_{change}$ are not necessarily SMLs with respect to any Invariance property $P$ because for all of these operators $\uparrow$ G.*

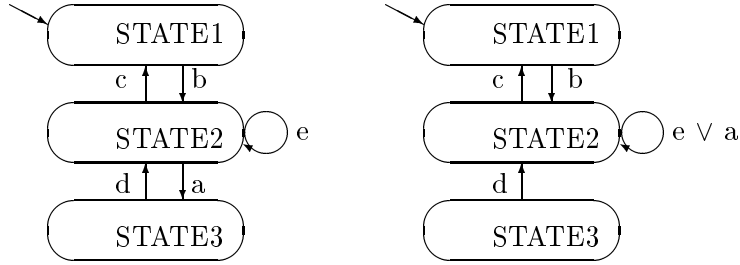

Figure 9: The automata S1 (left) and S1′ (right).

The next theorem characterizes those learning operators that cannot be guaranteed to be SMLs with respect to Response properties.

**Theorem 4** *Any machine learning operator schema that can introduce a new string with an infinitely repeating substring into the FSA language cannot be guaranteed to be an SML for Response properties.*

**Proof.** Assume FSA $S$ satisfies a Response property prior to learning. Therefore every string accepted by $S$ satisfies the property. For each accepted string, every instance (or the first instance if it is a First-Response property) of the trigger is eventually followed by a response. Suppose the machine learning operator introduces a new string with an infinitely repeating substring into the language. Then it is possible that the prefix of this string before the infinitely repeating substring includes a trigger and no response, and the infinitely repeating substring does not include a response. □

Since we already have results to cover the one-step operators, we need only consider the two-step operators.

**Corollary 7** *All of the two-step learning operators cannot be guaranteed to be SMLs with respect to Response properties because they are in the first class in the partition related to this theorem, i.e., they may introduce strings with infinitely repeating substrings.*

Consider a couple of illustrative examples of Theorem 4 and its corollary, using Figure 9. Prior to learning (the FSA on the left of Figure 9), $\forall \mathbf{x}$, where $\mathbf{x} \in \mathcal{L}(S1)$, $\mathbf{x} \models$ P3, for Response property P3 = □ (a → ◇ d). Assume operator $o_{stay}$: S1 → S1′ deletes edge (STATE2, STATE3) and generalizes the transition condition on edge (STATE2, STATE2) to "e ∨ a" (see Figure 9 on the right). Then the string consisting of b followed by infinitely many a's (b,a,a,a,...) ∈ $\mathcal{L}(S1′)$ but $\not\models$ P3. This helps us to see why $o_{stay}$ is not necessarily an SML for Response properties. The same example illustrates why $o_{delete+gen}$ cannot be guaranteed to be an SML for Response properties. For $o_{spec+gen}$, suppose the condition for (STATE2, STATE3) is "f ∨ a" in S1, and "f ∧¬ a" in S1′ but everything else is the same as in Figure 9. Again, we can see the problem for Response properties.

We conclude by summarizing the positive a priori results:

- $o_{delete}$, $o_{spec}$, $o_{delete \vee spec}$ and $o_{delete-action}$ are SMLs for any property (expressible in temporal logic).

- $o_{delete+gen}$, $o_{spec+gen}$ and $o_{stay}$ are SMLs for Invariance properties.

and the negative a priori results:

- $o_{add}$, $o_{gen}$, $o_{add \lor gen}$, $o_{add-action}$, $o_{spec+add}$, $o_{delete+add}$, $o_{move}$ and $o_{change}$ are not necessarily SMLs for Invariance or Response properties.

- $o_{delete+gen}$, $o_{spec+gen}$ and $o_{stay}$ are not necessarily SMLs for Response properties.

The fact that all three learning operators that satisfy the accessibility condition are guaranteed to be SMLs for Invariance properties is significant, because Invariance properties are extremely useful and common for verifying systems and many important applications need only test properties of this class (Heitmeyer et al., 1998).

Finally, from Theorems 1 and 2 we learned that the heart of the problem for all of the negative results is either an $o_{gen}$ step or an $o_{add}$ step. Later in this paper we address these troublesome steps by finding more efficient methods for dealing with them than total reverification from scratch. However, first, in the next section, we consider how our a priori results are translated from a single to a product FSA for $SIT_{multplans}$.

## 5. Translating Learning Operators to a Product Automaton

In this section we address $SIT_{multplans}$ where each agent maintains and uses its own individual FSA, but for verification the product FSA needs to be formed and verified. For $SIT_{multplans}$, a learning operator is applied to an individual agent FSA and then the product is formed. Therefore, it is necessary to consider the translation of each learning operator from individual to product FSA, and how that affects the a priori SML results presented above.

For operators $o_{spec+gen}$, $o_{delete+gen}$, $o_{spec+add}$, and $o_{delete+gen}$, we consider only the translations of the primitive operators. This is because the translations of these operators are simply translations of their primitive components. The remaining translations are:

- $o_{spec}$ translates to $o_{spec}$ and/or $o_{delete}$.

- $o_{delete}$ translates to $o_{spec}$ and/or $o_{delete}$.

- $o_{gen}$ translates to $o_{gen}$ and/or $o_{add}$.

- $o_{add}$ translates to $o_{gen}$ and/or $o_{add}$.

- $o_{stay}$ translates to $o_{stay}$ and/or $o_{move}$.

- $o_{move}$ translates to $o_{move}$.

- $o_{change}$ translates to $o_{change}$.

It may not be intuitive to the reader how $o_{gen}$ can translate to $o_{add}$. To illustrate, we use Figure 10, where the transition conditions, such as (a $\lor$ c), denote sets of *multiagent* actions. Suppose $o_{gen}$ is applied to edge (1, 2) in the leftmost FSA so that the transition condition is now (d $\lor$ b). Then a new edge $(11', 21')$ is added to the product FSA (rightmost in Figure 10) with the transition condition b. Recall that to form the product FSA we take the Cartesian product of the vertices and the intersection of the transition conditions. Likewise, $o_{spec}$ translates to either $o_{spec}$ or $o_{delete}$ in the product FSA.

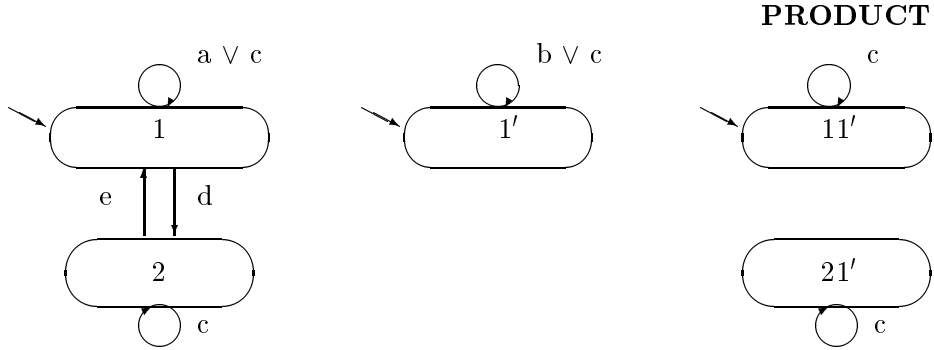

Figure 10: Generalization can become addition in product.

To illustrate why $o_{stay}$ can become $o_{move}$ in the product, we use Figure 3. Suppose we delete the edge (TRANSMITTING, RECEIVING) and move the transition condition to edge (TRANSMITTING, TRANSMITTING). Then the global state (DELIVERING, DELIVERING, TRANSMITTING) becomes accessible from initial state (COLLECTING, RECEIVING, TRANSMITTING) by taking multiagent action (F-deliver $\wedge$ I-receive $\wedge$ L-transmit). Previously, that multiagent action forced the product FSA to go to (DELIVERING, DELIVERING, RECEIVING).

What implications do these translations have for the safety of the learning operators for the product FSA? The positive a priori results for $o_{delete+gen}$, $o_{spec+gen}$, and $o_{stay}$ for preserving Invariance properties become negative for the product. This is because $o_{gen}$ may become $o_{add}$ and $o_{stay}$ may become $o_{move}$. On the other hand, the positive a priori results for $o_{delete}$, $o_{spec}$, $o_{delete \vee spec}$ and $o_{delete-action}$ preserving all properties remain positive for the product. For $o_{delete}$, $o_{spec}$, $o_{delete \vee spec}$, and $o_{delete-action}$, this implies that the product FSA never needs to be formed, reverification does not have to be done, and thus there is *no* run-time cost, even for multiple agents learning autonomously. As mentioned above, the troublesome parts of all operators are due to their $o_{gen}$ or $o_{add}$ component. In the next section we develop methods for reducing the complexity of reverification over total reverification from scratch when these operators have been applied.

## 6. Incremental Reverification

Recall that operators $o_{spec}$ and $o_{delete}$ cannot cause problems with the safety of learning, whereas $o_{gen}$ and $o_{add}$ are risky (i.e., are not a priori guaranteed to be SMLs). Furthermore, $o_{gen}$ and $o_{add}$ can cause problems when they are the second step in a two-step operator. Fortunately, we have developed incremental reverification algorithms for these operators that can significantly decrease the time complexity over total reverification from scratch.

Recall that there are two ways that operators can alter accessibility: globally (G) or locally (L). Furthermore, recall that $o_{add}$ can increase accessibility either way ($\uparrow$ G $\uparrow$ L), whereas $o_{gen}$ can only increase accessibility locally ($\not\uparrow$ G $\uparrow$ L). We say that $o_{gen}$ has only a "localized" effect on accessibility, whereas the effects of $o_{add}$ may ripple through many parts of the FSA. The implication is that we can have very efficient incremental methods for reverification tailored for $o_{gen}$, whereas we cannot do likewise for $o_{add}$. In other words, a more localized effect on accessibility implies that it is easier to localize reverification to gain speed. This is also true for both two-step operators that have $o_{gen}$ as their second step, i.e., $o_{delete+gen}$ and $o_{spec+gen}$ are amenable to incremental (localized) reverification. Because no advantage is gained by considering $o_{add}$ per se, we develop incremental reverification algorithms for the most general operator $o_{change}$. These algorithms apply to $o_{add}$ and all other special cases of $o_{change}$.

We have developed two types of incremental reverification algorithms: those that follow the application of $o_{gen}$, and those that follow the application of $o_{change}$. For each of our learning operators, one or more of these algorithms is applicable. Before presenting the incremental algorithms, Subsection 6.1 presents two algorithms for total reverification from scratch, namely, one for Invariance properties and the other for all properties expressible as FSAs, as well as an algorithm for taking the tensor product of the FSAs. These algorithms apply to $SIT_{1agent}$, $SIT_{1plan}$, or $SIT_{multplans}$. Subsection 6.2 gives incremental versions of all the algorithms in Subsection 6.1. These algorithms are applicable when the learning operator is $o_{change}$ or any of its special cases. Furthermore, they apply to any of $SIT_{1agent}$, $SIT_{1plan}$, or $SIT_{multplans}$. Subsection 6.3 has incremental algorithms for $SIT_{1agent}$ and $SIT_{1plan}$, learning operator $o_{gen}$, and Invariance and full Response properties in particular. The section concludes with theoretical and empirical results comparing the time complexity of the incremental algorithms with the time complexity of the corresponding total version (as well as with each other).

The goal in developing all of the incremental reverification algorithms is maximal efficiency. These algorithms make the assumption that $S \models P$ prior to learning, which means that any errors found on previous verification(s) have already been fixed. Then learning occurs ($o(S) = S'$), followed by the incremental reverification algorithm (see Figure 1). Next let us consider the *soundness* and *completeness* of the algorithms, where we assume normal termination. All of the incremental reverification algorithms presented here are sound (i.e., whenever they conclude that reverification succeeds, it is in fact true that $S' \models P$) for "downstream" properties and "directionless" properties for which the negation is expressible as a Büchi FSA. Downstream properties (which include Response) check sequences of events in temporal order, e.g., whether every $p$ is followed by a $q$. In contrast, "upstream" properties check for events in reverse temporal order, e.g., whether every $q$ is preceded by a $p$.[10] Directionless properties, such as Invariance, impose no order for checking. Some of the incremental algorithms are also complete, i.e., whenever they conclude that reverification fails, it is in fact true that $S' \not\models P$. (The reader should avoid confusing "complete algorithm" with "complete FSA.")

When reverification fails, it does so because of one or more errors, where an "error" implies there is a property violation ($S' \not\models P$). There are two ways to resolve such errors. Either return to the prelearning FSA(s) and choose another learning operator and reverify again, or keep the results of learning but repair the FSA(s) in some other way to fix the error. With one exception, the complete algorithms in this section find *all* true errors introduced by learning. The algorithms that are not complete may also find false errors. Any algorithm that finds all and only true errors can resolve these errors in either of the two ways. An algorithm that does not find all errors or finds false ones requires more restricted error resolution. In particular, it can only be used with the first method for resolving errors, which consists of choosing another learning operator. The algorithms that are sound but not complete (can find false errors) are overly cautious. In other words, they may recommend avoiding a learning operator when in fact the operator may be safe to apply.

---

10. William Spears (personal communication) identified the upstream/downstream distinction as being relevant to the applicability of the incremental algorithms described here.

| | crt | crr | crp | cdt | cdr | cdp | drt | drr | drp | ddt | ddr | ddp |
|---|---|---|---|---|---|---|---|---|---|---|---|---|
| T | R | 0 | P | T | 0 | T | R | 0 | P | T | 0 | T |
| R | 0 | T | 0 | 0 | R | 0 | 0 | T | 0 | 0 | R | 0 |
| P | 0 | 0 | R | 0 | 0 | T | 0 | 0 | R | 0 | 0 | T |

Table 1: The transition function for agent L's FSA plan. The rows correspond to states and the columns correspond to multiagent actions.

Before presenting the incremental algorithms, we first present algorithms for total verification from scratch. These algorithms do not assume that learning has occurred, and they apply to all situations. They are more general (not tailored for learning), but less efficient, than our incremental algorithms.

## 6.1 Product and Total (Re)verification Algorithms for All Situations

For implementation efficiency, all of our algorithms assume that FSAs are represented using a table of the transition function $\delta(v_1, a) = v_2$, which means that for state $v_1$, taking action $a$ leads to next state $v_2$, as shown in Table 1. Rows correspond to states and columns correspond to multiagent actions. This representation is equivalent to the more visually intuitive representation of Figures 2 and 3. In particular, Table 1 is equivalent to the FSA in Figure 3 for the lander agent L. In Table 1, states are abbreviated by their first letter, and the multiagent actions are abbreviated by their first letters. For example, "crt" means agent F takes action (F-collect), I takes (I-receive), and L takes (L-transmit). The table consists of entries for the next state, i.e., it corresponds to the transition function. A "0" in the table means that there is no possible transition for this state-action pair. One situation in which this occurs is when an action is not allowed from a state. Consider an example use of the table format for finite-state automata. According to the first (upper leftmost) entry in Table 1, if L is in state TRANSMITTING ("T") and F takes action F-collect, I takes I-receive, and L takes L-transmit (which together is multiagent action "crt"), then L will transition to its RECEIVING ("R") state, i.e., $\delta(T, \text{crt}) = R$. With this tabular representation, $o_{change}$ is implemented as a perturbation (mutation) operator that changes a table entry to another randomly chosen value for the next state. Operator $o_{gen}$ is a perturbation operator that changes a 0 entry to a next state already appearing in that row. For example, generalizing the transition condition along edge (T,R) can be accomplished by changing one of the 0s to an R in the first row of Table 1. This is because the transition condition associated with edge (T,R) is the set of all multiagent actions that transition from Suppose there are $n$ agents, and $1 \leq j_k \leq$ the number of states in the FSA for agent $k$. Then the algorithm forms all product states $v = (v_{j_1}, .., v_{j_n})$ and specifies their transitions:

```
Procedure product
for each product state v = (v_{j_1}, .., v_{j_n}) do
        if all v_{j_k}, 1 ≤ k ≤ n, are initial states, then v is a product initial state
        endif
        for each multiagent action a_i do
                if (δ(v_{j_k}, a_i) == 0) for some k, 1 ≤ k ≤ n, then δ(v, a_i) = 0
                else δ(v, a_i) = (δ(v_{j_1}, a_i),...., δ(v_{j_n}, a_i)); endif
        endfor
endfor
end procedure
```

Figure 11: $Total_{prod}$ product algorithm.

T to R, i.e., {crt, drt} in Table 1. This set is expressed in Boolean algebra as (I-receive $\wedge$ L-transmit) (see Figure 3).

For $SIT_{1plan}$ or $SIT_{multplans}$, prior to verification the multiagent product FSA $S$ needs to be formed from the individual agent FSAs (see Figure 1). We can implement the algorithm $Total_{prod}$ for generating the product FSA using the data structure of Table 1 as shown in Figure 11. In the product FSA, an example product state and transition is $\delta$(CRT, drt) = DDR because $\delta$(C, drt) = D, $\delta$(R, drt) = D, and $\delta$(T, drt) = R for agents F, I, and L, respectively. The initial states of the product FSA are formed by testing whether every individual state of the product is an initial state. For example, if D, D, and R are initial states for F, I, and L, respectively, then DDR will be an initial state for F $\otimes$ I $\otimes$ L. After forming the product states and specifying which are initial, the algorithm of Figure 11 specifies the $\delta$ transition for every product state and multiagent action.

Note that the algorithm in Figure 11 forms the product FSA $S$ for testing Invariance properties. To test First-Response properties using AT verification, we need to form the product FSA $S \otimes \neg P$. To do this simply requires considering $\neg P$ to be the $(n+1)$st agent. The algorithm in Figure 11 is modified by changing $n$ to $n+1$ everywhere. It is also important to note that in *all* situations (including $SIT_{1agent}$), $Total_{prod}$ must be executed to form the product $S \otimes \neg P$ if AT verification is to be done. In $SIT_{1agent}$, $S$ is just the single agent FSA and $n$ is 1. For $SIT_{1plan}$, $n = 1$ also. In other words, for $SIT_{1plan}$ the multiagent plan, once formed, is never subdivided and therefore it could be considered like a single agent plan. In both of these cases, if AT verification is done the product is taken of the single plan FSA and the property FSA.

Given that the product FSA has been formed if needed, then the final (multi)agent FSA can be verified. We first consider a very simple model checking algorithm, called $Total_I$, tailored specifically for verifying Invariance properties of the form $\square \neg p$. The algorithm, shown in Figure 12, consists of a depth-first search of $S$ beginning in each initial state. Any accessible atom $a_i$ that is part of a transition condition, where $a_i \preceq p$, violates the property. (We store the set of all atoms $a_i \preceq p$ for rapid access.)

```
Procedure verify
      for each state v ∈ V(S) do
            visited(v) = 0
      endfor
      for each initial state v ∈ I(S) do
            if (visited(v) == 0) then dfs(v); endif
      endfor
end procedure
Procedure dfs(v)
      visited(v) = 1;
      for each atom a_i ∈ Γ(K), a_i ≼ p, do
            if δ(v, a_i) ≠ 0 then print "Verification error"; endif
      endfor
      for each atom a_i ∈ Γ(K) set w = δ(v, a_i) and do
            if (w ≠ 0) and (visited(w) == 0) then dfs(w); endif
      endfor
end procedure
```

Figure 12: $Total_I$ verification algorithm.

Next we consider an algorithm for verifying *any* property whose negation is expressible as a Büchi FSA, including First-Response properties. The reader may wish to review the high-level description of this AT model checking algorithm presented in Subsection 3.3 before continuing. Figure 13 gives a basic version of this algorithm from Courcoubetis et al. (1992) and Holzmann et al. (1996).[11] We call this algorithm $Total_{AT}$ because it is total automata-theoretic verification. Recall that in AT model checking, the property is represented as an FSA, and asking whether $S \models P$ is equivalent to asking whether $\mathcal{L}(S) \subseteq \mathcal{L}(P)$ for property $P$. This is equivalent to $\mathcal{L}(S) \cap \overline{\mathcal{L}(P)} = \emptyset$, which is algorithmically tested by taking the tensor product of the plan FSA and the FSA corresponding to $\neg P$. If $\mathcal{L}(S \otimes \neg P) = \emptyset$ then $\mathcal{L}(S) \subseteq \mathcal{L}(P)$, i.e., $S \models P$ and verification succeeds; otherwise, $S \not\models P$ and verification fails. The algorithm of Figure 13 assumes that the negation of the property ($\neg P$) is expressed as a Büchi automaton and the FSA being searched is $S \otimes \neg P$.

Algorithm $Total_{AT}$, in Figure 13, actually checks whether $S \not\models P$ for any property $P$. To check if $S \not\models P$, we can determine whether $\mathcal{L}(S \otimes \neg P) \neq \emptyset$. This is true if there is some "bad" state in $B(S \otimes \neg P)$ reachable from an initial state and reachable from itself, i.e., part of an accessible cycle and therefore visited infinitely often. The algorithm of Figure 13 performs this check using a nested depth-first search on the product FSA $S \otimes \neg P$. The first depth-first search begins at initial states and visits all accessible states. Whenever a state $s \in B(S \otimes \neg P)$ is discovered, it is called a "seed," and a nested search begins to look for a cycle that returns to the seed. If there is a cycle, this implies the $B(S \otimes \neg P)$ (seed) state can be visited infinitely often, and therefore the language is nonempty (i.e., there is some action sequence in the plan that does not satisfy the property) and verification fails.

---

11. This algorithm is used in the well-known Spin system (Holzmann, 1991). A modification was made to the published algorithm for readability, as well as for efficiency, for the case where it is desirable to halt after the first verification error. This modification makes the nested call first in procedure dfs.

```
Procedure verify
       for each state v ∈ V(S ⊗ ¬P) do
              visited(v) = 0
       endfor
       for each initial state v ∈ I(S ⊗ ¬P) do
              if (visited(v) == 0) then dfs(v); endif
       endfor
end procedure
Procedure dfs(v)
       visited(v) = 1;
       if v ∈ B(S ⊗ ¬P) then
              seed = v;
              for each state v ∈ V(S ⊗ ¬P) do
                     visited2(v) = 0
              endfor
              ndfs(v)
       endif
       for each successor (i.e., next state) w of v do
              if (visited(w) == 0) then dfs(w); endif
       endfor
end procedure
Procedure ndfs(v)  /* the nested search */
       visited2(v) = 1;
       for each successor (i.e., next state) w of v do
              if (w == seed) then print "Bad cycle. Verification error";
              break
              else if (visited2(w) == 0) then ndfs(w); endif
              endif
       endfor
end procedure
```

Figure 13: $Total_{AT}$ verification algorithm.

Suppose there are $n$ agents, and agent $i$ was modified, $1 \leq i \leq n$.
Operator $o_{change}$ modified $\delta(v_i, a_{adapt})$ to be $w_i{}'$ for state $v_i$ and multiagent action $a_{adapt}$.
$1 \leq j_k \leq$ the number of states in the FSA for agent $k$.
Then the algorithm is:

```
Procedure product
for each product state v = (v_{j_1},...,v_i,...,v_{j_n}) formed from state v_i do
    if (δ(v_{j_k}, a_{adapt}) == 0) for some k, 1 ≤ k ≤ n, then δ(v, a_{adapt}) = 0
    else δ(v, a_{adapt}) = (δ(v_{j_1}, a_{adapt}),..., w_i', ..., δ(v_{j_n}, a_{adapt})); endif
endfor
end procedure
```

Figure 14: $Inc_{prod}$ product algorithm.

$Total_I$ and $Total_{AT}$ are sound and complete (for *any* property whose negation is expressible as a Büchi FSA), and they find all verification errors. Before elaborating on this, first note that the term "verification error" has a different connotation for $Total_I$ and $Total_{AT}$. For $Total_I$ an error is an accessible bad atom (i.e., an atom $a \preceq p$ where the property is $\Box \neg p$). For $Total_{AT}$ it is an accessible bad state that is part of a cycle. The reason $Total_I$ is sound is that it flags as errors only those atoms $a \preceq p$. It is complete and finds all errors because it does exhaustive search and testing of all accessible atoms. $Total_{AT}$ is also sound and complete, for analogous reasons. Because $Total_I$ and $Total_{AT}$ find all errors, they can be used with either method of error resolution (i.e., choose another operator or fix the FSA).

## 6.2 Incremental Algorithms for $o_{change}$ and All Situations

All of the algorithms in the previous subsection can be streamlined given that it is known that a learning operator (in this case, $o_{change}$) has been applied. For simplicity, all of our algorithms assume $o_{change}$ is applied to a single atom (multiagent action). For example, we assume that if $\delta(v_i, a_{adapt}) = w_i$, then $o_{change}(\delta(v_i, a_{adapt})) = w_i{}'$ where $w_i$ and $w_i{}'$ are states (or 0, implying no next state), and $a_{adapt}$ is a multiagent action. Since we use the tabular representation, this translates to changing one table entry.

Figure 14 shows an incremental version of $Total_{prod}$, called $Inc_{prod}$, which is tailored for re-forming the product FSA after $o_{change}$ has been applied. The algorithm of Figure 14 is for Invariance properties; for AT verification change $n$ to $n+1$ in the algorithm and assume $\neg P$ is the $(n+1)$st agent. Although $Inc_{prod}$ is applicable in all situations when taking the product with the property FSA, the primary motivation for developing this algorithm was the multiagent $SIT_{multplans}$. Recall that in this situation, every time learning is applied to an individual agent FSA, the product must be *re*-formed to verify global properties. The wasted cost of doing this motivated the development of this algorithm.

Algorithm $Inc_{prod}$ assumes that the product was formed originally (before learning) using $Total_{prod}$. $Inc_{prod}$ capitalizes on the knowledge of what (individual or multiagent) state ($v_i$) and multiagent action ($a_{adapt}$) transition to a new next state as specified by operator $o_{change}$. This algorithm assumes that the prelearning product FSA is stored. Then the only product FSA states whose next state needs to be modified are those states that

```
Procedure product
I(S) = ∅;
for each product state v = (v_{j_1}, ..., v_i, ..., v_{j_n}) formed from state v_i do
        if visited(v) then I(S) = I(S) ∪ {v}; endif
        if (δ(v_{j_k}, a_{adapt}) == 0) for some k, 1 ≤ k ≤ n, then δ(v, a_{adapt}) = 0
        else δ(v, a_{adapt}) = (δ(v_{j_1}, a_{adapt}), ..., w_i', ..., δ(v_{j_n}, a_{adapt})); endif
endfor
end procedure
```

Figure 15: $Inc_{prod-NI}$ product algorithm: a variation of $Inc_{prod}$ that gets new initial states.

include $v_i$ and transition on action $a_{adapt}$. The method for reverification that is assumed to follow $Inc_{prod}$ is total reverification, i.e., $Total_I$ or $Total_{AT}$.

Next, consider another pair of product and reverification algorithms that is expected to be, overall, potentially even more efficient. The goal is to streamline reverification after $o_{change}$. This requires a few simple changes to the algorithms. The motivation for these changes is that when model checking downstream properties, $o_{change}$ has only "downstream effects," i.e., it only affects the accessibility of vertices and atoms altered by $o_{change}$ or those that would be visited by verification *after* those altered by $o_{change}$.

Consider the changes. We start by building a set of the Cartesian product states $v = (v_{j_1}, ..., v_i, ..., v_{j_n})$ that are formed from the state $v_i$ that was affected by learning. The first way that we can shorten reverification is by using these states as the new initial states for reverification. In fact, we need only select those that were visited during the original verification (i.e., are accessible from the original initial states). In other words, suppose for agent $i$, $o_{change}$ modified $δ(v_i, a_{adapt})$. Then we reinitialize the set of initial states to be $∅$ and add all product states formed from $v_i$ that were marked "visited" during previous verification. This can be done by modifying the product algorithm of Figure 14 as shown in Figure 15. The algorithm of Figure 15 is to form the product FSA for verifying Invariance properties. To form the product for AT verification, substitute $I(S ⊗ ¬P)$ for $I(S)$ and $(n + 1)$ for $n$ in Figure 15. We call this incremental product algorithm $Inc_{prod-NI}$, where "NI" denotes the fact that we are getting new initial states.

The second way to streamline reverification is by only considering a transition on action $a_{adapt}$, the action whose $δ$ value was modified by learning, from these new initial states. Thereafter, incremental reverification proceeds exactly like total (re)verification. With these changes, $Total_I$ becomes $Inc_{I-NI}$, shown in Figure 16. Likewise, with these changes $Total_{AT}$ becomes $Inc_{AT-NI}$, as shown in Figure 17. Figure 17 shows only changes to procedures dfs and verify; ndfs is the same as in Figure 13. One final streamlining added to $Inc_{I-NI}$, but not $Inc_{AT-NI}$, is that only the new initial states have "visited" reinitialized to 0. This can be done for Invariance properties because they are not concerned with the order of atoms in strings.[12]

---

12. Suppose $o_{change}$ adds a new edge $(v_1, v_3)$. If $v_3$ was visited on previous verification of an Invariance property (from a state other than $v_1$), then all atoms that can be visited after $v_3$ would already have been tested for the property. On the other hand, when testing First-Response properties the order of

```
Procedure verify
        for each new initial state v ∈ I(S) do
                visited(v) = 0
        endfor
        for each new initial state v ∈ I(S) do
                if (visited(v) == 0) then dfs(v); endif
        endfor
end procedure
Procedure dfs(v)
        visited(v) = 1;
        if v ∈ I(S) and w ≠ 0, where w = δ(v, a_adapt), then
                if (a_adapt ⪯ p) then print "Verification error"; endif
                if (visited(w) == 0) then dfs(w); endif
        else
                for each atom a_i ∈ Γ(K), a_i ⪯ p, do
                        if δ(v, a_i) ≠ 0 then print "Verification error"; endif
                endfor
                for each atom a_i ∈ Γ(K) set w = δ(v, a_i) and do
                        if (w ≠ 0) and (visited(w) == 0) then dfs(w); endif
                endfor
        endif
end procedure
```

Figure 16: $Inc_{I-NI}$ reverification algorithm.

$Inc_{I-NI}$ is sound for Invariance properties, and $Inc_{AT-NI}$ is sound for any downstream or directionless property whose negation is expressible as a Büchi FSA, including First-Response and Invariance. Assuming $S \models P$ prior to applying $o_{change}$ to form $S'$, if these incremental reverification algorithms conclude that $S' \models P$, then total reverification would also conclude that $S' \models P$. Recall that total reverification is sound. Therefore, the same is true for these incremental algorithms. Furthermore, these incremental reverification algorithms will find all of the new violations of the property introduced by $o_{change}$. The reason the algorithms are sound and find all new errors (for downstream or directionless properties) is that there are only two ways that accessibility can be modified by any of our learning operators, including $o_{change}$: locally or globally. Recall that local change alters the accessibility of atom $a_{adapt}$ or the state $\delta(v_i, a_{adapt})$, and a global change alters the accessibility of states or atoms that would be visited *after* $\delta(v_i, a_{adapt})$. In neither case (local or global) will the learning operator modify accessibility of atoms or states visited before, but not after, $a_{adapt}$. Our algorithms reverify exhaustively (i.e., they reverify as much as total reverification would) for all atoms and states visited at or after $a_{adapt}$. Since these incremental algorithms perform reverification exactly the same way as their total versions

---

atoms is relevant. Even if $v_3$ was previously visited, since it might not have been visited from $v_1$, the addition of $(v_1, v_3)$ could add a new string with a new atom order that might violate the First-Response property. Therefore, $v_3$ needs to be revisited for First-Response properties, but not for Invariance properties.

```
Procedure verify
      for each state v ∈ V(S ⊗ ¬P) do
            visited(v) = 0
      endfor
      for each new initial state v ∈ I(S ⊗ ¬P) do
            if (visited(v) == 0) then dfs(v); endif
      endfor
end procedure
Procedure dfs(v)
      visited(v) = 1;
      if v ∈ B(S ⊗ ¬P) then
            seed = v;
            for each state v ∈ V(S ⊗ ¬P) do
                  visited2(v) = 0
            endfor
            ndfs(v)
      endif
      if v ∈ I(S ⊗ ¬P) and w ≠ 0 and (visited(w) == 0),
      where w = δ(v, a_adapt), then dfs(w)
      else
            for each successor (i.e., next state) w of v do
                  if (visited(w) == 0) then dfs(w); endif
            endfor
      endif
end procedure
```

Figure 17: Procedures verify and dfs of the $Inc_{AT-NI}$ reverification algorithm.

do after the part of the FSA that was modified by learning, they will find all new errors introduced by learning.

$Inc_{I-NI}$ is complete for Invariance properties because it flags errors using the same method as $Total_I$, and because Invariance properties are directionless and are therefore impervious to the location of atoms in a string. On the other hand, $Inc_{AT-NI}$ is not complete for all downstream properties. For example, it is not complete for properties that check for the $first$ occurrence of a pattern in a string, e.g., First-Response properties. Because $Inc_{AT-NI}$ does not identify whether the new initial states are before or after the first occurrence, there is no way to know if the first occurrence is being checked after learning. Nevertheless, this lack of completeness for First-Response properties actually turns out to be a very useful trait, as we will discover in Subsection 6.5.

## 6.3 Incremental Algorithms for $o_{gen}$ and $SIT_{1agent/1plan}$

We next present our final two incremental reverification algorithms, which are applicable only in $SIT_{1agent}$ and $SIT_{1plan}$, when there is one FSA to reverify. These are powerful algorithms in terms of their capability to reduce the complexity of reverification. However, their soundness relies on the assumption that the learning operator's effect on accessibility

```
Procedure check-invariance-property
if v₁ was not previously visited, then output "Verification succeeds."
else
        if (z ⊨ ¬p) then output "Verification succeeds."
        else output "Avoid this instance of o_gen."; endif
endif
end procedure
```

Figure 18: $Inc_{gen-I}$ reverification algorithm.

is localized, i.e., that it is $o_{gen}$ with $SIT_{1agent}$ or $SIT_{1plan}$ but not $SIT_{multplans}$ (where $o_{gen}$ might become $o_{add}$). An important advantage of these algorithms is that they never require forming a product FSA, not even $S \otimes \neg P$, regardless of whether the property is type Response. The algorithms gain efficiency by being *both* tailored to a specific property type *and* to a specific learning operator. The objective in developing these algorithms was maximal efficiency and therefore they sacrifice completeness and/or the ability to find all errors.

These two incremental algorithms are tailored for reverification after operator $o_{gen}$. Assume that property $P$ holds for $S$ prior to learning, i.e., $S \models P$. Now we generalize the transition condition $M_{\mathcal{K}}(v_1, v_3) = y$ to form $S'$ via $o_{gen}$ $(M_{\mathcal{K}}(v_1, v_3)) = y \vee z$, where $y \wedge z = 0$. We want to verify that $S' \models P$.

One additional definition is needed before presenting our algorithms. We previously defined what it means for a c-state formula $p$ to be true at a c-state, but to simplify the algorithms we also define what it means for a c-state formula to be true of a transition condition. A c-state formula $p$ is defined to be true of a transition condition $y$, i.e., "$y \models p$," if and only if $y \preceq p$ (which can be implemented by testing whether for every atom $a \preceq y$, $a \preceq p$.)

Let us begin with the algorithm $Inc_{gen-I}$ (which consists of two very simple tests) tailored for $o_{gen}$ and Invariance properties, shown in Figure 18. Recall that $M_{\mathcal{K}}(v_1, v_3) = y$ and $o_{gen}(M_{\mathcal{K}}(v_1, v_3)) = y \vee z$. $Inc_{gen-I}$, which tests "$z \models \neg p$," localizes reverification to a restricted portion of the FSA. (For efficiency, $z \models \neg p$ is implemented as a test for $z \preceq p$ rather than $z \preceq \neg p$ because $p$ is typically expected to be more succinct than $\neg p$.) Assume the Invariance property is $P = \Box \neg p$ and $S \models P$. Then every string $\mathbf{x}$ in $\mathcal{L}(S)$ satisfies Invariance property $P$, so for each $\mathbf{x}$, $\neg p$ is true of every atom in $\mathbf{x}$. This implies $y \models \neg p$. This statement is based on our assumption that $v_1$ is accessible from an initial state. If not, reverification is not needed. The generalization will not violate $P$. Therefore, the algorithm begins by testing whether $v_1$ was visited on previous verification. If not, the output is "success." (Note that $o_{gen}$ does not alter the accessibility of $v_1$.)

$Inc_{gen-I}$ is sound and complete for Invariance properties. Generalization of $M_{\mathcal{K}}(v_1, v_3)$ is application of $o_{gen}$ $(M_{\mathcal{K}}(v_1, v_3)) = y \vee z$ to form $S'$. This operator $o_{gen}$ preserves Invariance property $P$ if and only if $S' \models P$, which is true if and only if $z \models \neg p$. The reason for this is that we know $S$ satisfies $P$ from our original verification, and therefore $\neg p$ is true for all atoms in all strings in $\mathcal{L}(S)$. The only possible new atoms in $\mathcal{L}(S')$ but not in $\mathcal{L}(S)$ are in $z$. If $z \models \neg p$, then $\neg p$ is true for all atoms in $\mathcal{L}(S')$, which implies that every string in

```
Procedure check-response-property
if y ⊨ q then
        if (z ⊨ q and z ⊨ ¬p) then output "Verification succeeds."
        else output "Avoid this instance of o_gen"; endif
else
        if (z ⊨ ¬p) then output "Verification succeeds."
        else output "Avoid this instance of o_gen"; endif
endif
end procedure
```

Figure 19: $Inc_{gen-R}$ reverification algorithm.

$\mathcal{L}(S')$ satisfies $P$. In other words, $S' \models P$. Therefore, $Inc_{gen-I}$ is sound. We also know that it is complete because if $\exists a, a \preceq z, a \npreceq p$, then it must be the case that $S' \not\models P$. In conclusion, $Inc_{gen-I}$, which consists of the test "$z \models \neg p$," is sound and complete. For maximal efficiency, our implementation of $Inc_{gen-I}$ halts after the first error, although it is simple to modify it to find all errors (and this does not significantly affect the empirical time complexity results of Subsection 6.5, nor does it affect the worst-case time complexity). $Inc_{gen-I}$ is incremental because it is localized to just checking whether the property holds of the newly added atoms in $z$, rather than all atoms in $\mathcal{L}(S')$. Finally, this algorithm only needs to be executed for $o_{gen}$, but not for $o_{spec+gen}$ or $o_{delete+gen}$, because $o_{gen}$ is the only version that can add *new* atoms via generalization. Recall that $o_{spec+gen}$ and $o_{delete+gen}$ are SMLs for Invariance properties.

As an example of $Inc_{gen-I}$, suppose a, b, c, d, and e are atoms, and the transition condition $y$ between STATE1 and STATE2 equals a. Let (a, b, b, d, d,...), where the ellipsis indicates infinite repetition of d, be a string in $\mathcal{L}(S)$ that includes STATE1 and STATE2 as the first two vertices in its accepting run. The property is $P = \Box \neg$ e. Assume the fact that this string satisfies $\neg$ e was proved in the original verification. Suppose $o_{gen}$ generalizes $M_{\mathcal{K}}$(STATE1, STATE2) from a to (a ∨ c) (i.e., it adds a new allowable action c from STATE1), which adds the string (c, b, b, d, d,...) to $\mathcal{L}(S')$. Then rather than test whether all of the elements of { a, b, c, d } are $\preceq \neg$ e, we really only need to test whether c $\preceq \neg$ e, because c is the only newly added atom.

The next algorithm, $Inc_{gen-R}$, is for generalization and full Response properties (and is nothing more than some simple tests). Like $Inc_{gen-I}$, $Inc_{gen-R}$ localizes reverification to a restricted portion of the FSA. Assume the Response property is $P = \Box(p \rightarrow \Diamond q)$, where $p$ is the trigger and $q$ is the response for c-state formulae $p$ and $q$. Assume property $P$ holds for $S$ prior to learning ($S \models P$). Now we generalize $M_{\mathcal{K}}(v_1, v_3) = y$ to form $S'$ by applying $o_{gen}$ $(M_{\mathcal{K}}(v_1, v_3)) = y \vee z$, where $y \wedge z = 0$. We need to verify that $S' \models P$.

$Inc_{gen-R}$ for $o_{gen}$ and full Response properties is in Figure 19. ($Inc_{gen-R}$ is also applicable for $o_{delete+gen}$ and $o_{spec+gen}$.) The algorithm first checks whether a response could be required of the transition condition $M_{\mathcal{K}}(v_1, v_3)$. A response is required if, for at least one string in $\mathcal{L}(S)$ whose run includes $(v_1, v_3)$, the prefix of this string before visiting vertex $v_1$ includes the trigger $p$ not followed by response $q$, and the string suffix after $v_3$ does not include $q$ either. Such a string satisfies the property if and only if $y \models q$. Thus if $y \models q$

135

and the property is true prior to learning (i.e., for $S$), then it is possible that a response is required. In this situation (i.e., $y \models q$), the only way to be sure we are safe ($S' \models P$) is if the newly added condition $z$ also has the response, i.e., $z \models q$. If not, then there could be new strings in $\mathcal{L}(S')$ whose accepting runs include $(v_1, v_3)$ but do not satisfy the property. For example, suppose a, b, c, and d are atoms, and the transition condition $y$ between STATE4 and STATE5 equals d. Let $\mathbf{x} = $ (a, b, b, d, ...) be a string in $\mathcal{L}(S)$ that includes STATE4 and STATE5 as the fourth and fifth vertices in its accepting run. The property is $P = \square$ (a $\rightarrow \diamond$ d), and therefore $y \models q$ and $\mathbf{x} \models P$. Suppose $o_{gen}$ generalizes $M_\mathcal{K}$(STATE4, STATE5) from d to (d $\vee$ c), where $z$ is c, which adds the string $\mathbf{x}' = $ (a, b, b, c, ...) to $\mathcal{L}(S')$. Then $z \not\models q$. If the string suffix after (a, b, b, c) does not include d, then there is now a string that includes the trigger but does not include the response. In other words, $\mathbf{x}' \not\models P$. Finally, if $y \models q$ and $z \models q$, an extra check is made to be sure $z \models \neg p$ because an atom could be both a response and a trigger. New triggers should be avoided.

The second part of the algorithm states that if $y \not\models q$ and no new triggers are introduced by generalization, then the operator is "safe" to do. It is guaranteed to be safe ($S' \models P$) in this case because if $y \not\models q$, then a response *cannot* be required here. In other words, because $S \models P$, for every string in $\mathcal{L}(S)$ whose accepting run includes $(v_1, v_3)$, either no trigger occurred prior to visiting $v_1$, or every trigger was followed by a response prior to visiting $v_1$, or a response occurred after visiting $v_3$.

$Inc_{gen-R}$ is sound but not complete for full Response properties. Its soundness is based on the fact that $o_{gen}$ does not increase accessibility of vertices or atoms visited after state $v_3$ (i.e., globally) and therefore reverification can be localized to only $M_\mathcal{K}(v_1, v_3)$. $Inc_{gen-R}$ is not complete because it may output "Avoid this instance of $o_{gen}$" when in fact $o_{gen}$ is safe to do. For example, if $y \models q$ but $z \not\models q$, the algorithm will output "Avoid this instance of $o_{gen}$." Yet it may be the case that $S' \models P$ if no trigger $p$ precedes response $q$ in $\mathcal{L}(S')$, or if a response is after $v_3$. When $Inc_{gen-R}$ outputs verification failure, it does not supply sufficient information for FSA repair. Errors must be resolved by selecting another learning operator. Note that "error" has a different connotation for $Inc_{gen-R}$ than for the AT verification algorithms. Any "Avoid..." output is considered an error.

Another disadvantage of $Inc_{gen-R}$ is that it does not allow generalizations that add triggers. If it is desirable to add new triggers during generalization, then one needs to modify $Inc_{gen-R}$ to call $Inc_{AT}$ when reverification with $Inc_{gen-R}$ fails, instead of outputting "Avoid this instance of $o_{gen}$." This modification also fixes the false error problem, *and* preserves the enormous time savings (see Section 6.5) when reverification succeeds.

## 6.4 Theoretical Worst-Case Time Complexity Analysis

Recall that one of our primary objectives is timely agent responses. This section compares the worst-case time complexity of the algorithms. Let us begin with the time complexity of $Total_{prod}$. This is $O((\prod_{i=1}^{n} |V(S_i)|) * |\Gamma(\mathcal{K})| * n)$ to form the product of the individual agent FSAs for Invariance property verification, and $O((\prod_{i=1}^{n} |V(S_i)|) * |P| * |\Gamma(\mathcal{K})| * n)$ to form the product for AT verification. Here $n$ is the number of agents, $|V(S_i)|$ is the number of states in single agent FSA $S_i$, $|P|$ is the number of states in the property FSA $P$, and $|\Gamma(\mathcal{K})|$ is the total number of atoms (multiagent actions). The reason for this complexity result is that there are $\prod_{i=1}^{n} |V(S_i)|$ product states for Invariance property verification, and $(\prod_{i=1}^{n} |V(S_i)|) * |P|$ product states for AT verification. The outer loop of $Total_{prod}$ iterates through all product states. The inner loop of $Total_{prod}$ iterates through all $|\Gamma(\mathcal{K})|$ atoms. Note that $(\prod_{i=1}^{n} |V(S_i)|) * |\Gamma(\mathcal{K})|$ and $(\prod_{i=1}^{n} |V(S_i)|) * |P| * |\Gamma(\mathcal{K})|$ are the sizes of the product FSA transition function tables built for $Total_I$ and $Total_{AT}$, respectively. $Total_{prod}$ does at most $n$ lookups for each table entry.

By comparison, our incremental algorithm $Inc_{prod}$ for generating the product FSA has time complexity $O((\prod_{i=1}^{n-1} |V(S_i)|) * n)$ or $O((\prod_{i=1}^{n-1} |V(S_i)|) * |P| * n)$ to modify the product FSA for Invariance property reverification or AT reverification, respectively. This is because the total number of revised product states is $(\prod_{i=1}^{n-1} |V(S_i)|)$ or $(\prod_{i=1}^{n-1} |V(S_i)|) * |P|$, and only one atom is considered (because we assume $o_{change}$ changes the next state for a single atom $a_{adapt}$). The time complexity of $Inc_{prod-NI}$ is the same as that of $Inc_{prod}$.

Next consider the worst-case time complexity of total (re)verification after the product has been formed. It is $O((\prod_{i=1}^{n} |V(S_i)|) * |\Gamma(\mathcal{K})|)$ for $Total_I$. This is because, in the worst case, every product state is accessible and therefore every entry in the product FSA transition function table is visited. Assuming $|B|$ is the number of "bad" (in the Büchi sense) states in the product FSA, then the worst-case time complexity of $Total_{AT}$ is $O((|B| + 1) * (\prod_{i=1}^{n} |V(S_i)|) * |P| * |\Gamma(\mathcal{K})|)$. This is because, in the worst case, every entry in the product FSA transition function table is visited once on the depth-first search and, for each bad state, again on the nested depth-first search. Unfortunately, the worst-case time complexity of $Inc_{I-NI}$ and $Inc_{AT-NI}$ are the same as that of $Total_I$ and $Total_{AT}$, respectively. This is because, in the worst case, every product state is still accessible. The restriction to transition only on $a_{adapt}$ at first does not reduce the "big O" complexity.

Finally, we consider the worst-case complexity of $Inc_{gen-I}$ and $Inc_{gen-R}$. First, we define for any Boolean expression $x$, $|x|$ is the number of elements in $\{a \mid a \in \Gamma(\mathcal{K}) \text{ and } a \preceq x\}$. For Invariance properties $P$ of the form $\Box \neg p$, $|P|$ equals $|p|$ since we test for each atom $a$ whether $a \models p$ rather than $a \models \neg p$, because we expect $|p| < |\neg p|$ in general. Then $Inc_{gen-I}$ requires time $O(|z| * |p|)$ to determine whether $z \models \neg p$. (Checking whether $v_1$ was visited requires constant time.) Assuming $|p| < \prod_{i=1}^{n} |V(S_i)|$ (which should be true except under bizarre circumstances), and since $|z| \leq \Gamma(\mathcal{K})$, $Inc_{gen-I}$ saves time over $Total_I$. $Inc_{gen-R}$ requires time $O((|y| * |q|) + (|z| * (|p| + |q|)))$ to determine whether $y \models q$, and then to determine whether $z \models q$ and $z \models \neg p$.[13] Clearly $(|y|+|z|) \leq |\Gamma(\mathcal{K})|$ because by the definition of $o_{gen}$, $y \wedge z = 0$. Therefore, assuming $(|p| + |q|) < ((|B| + 1) * |\prod_{i=1}^{n} |V(S_i)| * |P|)$ (which, again, should be true except in bizarre circumstances), the worst-case time complexity of $Inc_{gen-R}$ is lower than that of $Total_{AT}$.

## 6.5 Empirical Time Complexity Comparisons

Worst-case time complexity is not always a useful measure. Therefore we supplement the worst-case analyses with empirical results on cpu time. Our primary objective in these experiments is to compare the incremental algorithms with total reverification, as well as with each other, for the context of evolving behaviorally correct FSAs. The time required for reverification is significant to address if we want timely agent responses, because reverification occurs after *every* learning operator application.

---

13. Determining whether $z \models \neg p$ can be done by determining whether $z \preceq p$. Also, for $Inc_{gen-R}$, an additional time $O(|\Gamma(\mathcal{K})| - (|y| + |z|))$ is needed to identify $y$ when using the representation of Table 1. This does not affect our complexity comparisons or conclusions.

Before describing the experimental results, let us consider the experimental methodology. All code was written in C and run on a Sun Ultra 10 workstation. In our experiments, FSAs were randomly initialized, subject to certain restrictions. The reason for randomness is that this is a typical way to initialize individuals in a population for an evolutionary algorithm. There are two restrictions on the FSAs. First, although determinism and completeness of FSAs are execution, rather than verification, issues and therefore need not be enforced for these experiments, our choice of tabular representation of the FSAs (see Table 1) restricts the FSAs to being deterministic. Second, because the incremental algorithms assume $S \models P$ prior to learning, we restrict the FSAs to comply with this. There are two alternative methods for enforcing this in the experiments: (1) use sparse FSAs (i.e., with many 0s) and keep generating new FSAs until total verification succeeds (which does not take long with sparse FSAs), or (2) use dense FSAs engineered to guarantee property satisfaction. In particular, dense FSAs are forced to satisfy Invariance properties $\Box \neg p$ by inserting 0s in every column of the transition function table (such as Table 1) labeled with an atom $a \preceq p$. Dense FSAs are forced to satisfy First-Response properties with trigger $p$ and response $q$ by inserting 0s in every column labeled with an atom $a \preceq p$. This eliminates triggers initially. Note that either of these methods is a viable way to initialize a population of FSAs for evolution because it ensures early success in satisfying the property. This paper presents only the results with dense FSAs. See Gordon (1999) for the results with sparse FSAs.[14]

Another experimental design decision was to show scaleup in the size of the FSAs. Throughout the experiments there were assumed to be three agents, each with the same 12 multiagent actions. Each individual agent FSA had 25 or 45 states.[15] With 45 states the transition table contains $45^3 * 12$ entries.

A suite of five Invariance and five Response properties was used, which is in Appendix C. Invariance properties were expressed by storing the set of all atoms $a \preceq p$ for property $\Box \neg p$. This suffices for all of our algorithms tailored for Invariance properties. For AT verification, Response properties were expressed with a First-Response Büchi FSA for the negation of the property. An explanation of why this is adequate for our experiments is below. For $Inc_{gen-R}$, trigger $p$, and response $q$, all atoms $a_i \preceq p$ and $a_j \preceq q$ were stored. Six independent experiments were performed to verify each of the properties. In other words, every reverification algorithm was tested with 30 runs – six runs for each of five Invariance or five Response properties. For every one of these runs, a different random seed was used for generating the three FSAs. However, it is important to point out that all algorithms being compared with each other saw the *same* FSAs. For example, in Table 2 we compare $Inc_{prod}$ (row 1), $Inc_{prod-NI}$ (row 4), and $Total_{prod}$ (row 7). They all input the same three FSAs. Furthermore, the learning operator (specific instantiation of the operator schema) was the same for all algorithms being compared.

---

14. Sparse FSAs have an additional advantage, assuming they remain relatively sparse after evolution. The advantage is their succinctness for efficient execution, as in multientity models (Tennenholtz & Moses, 1989).

15. The sparse FSAs had 25, 45, or 65 states. To get accurate timing results with the dense FSAs, though, 65 states required a cpu free of any interfering processes for an unreasonably long time.

Let us consider the results in Tables 2 and 3. In both of these tables, each row corresponds to an algorithm. Rows are numbered for later reference. The entries give performance results, to be described shortly. Table 2 compares the performance of total reverification with the algorithms of Subsection 6.2, which were designed for $o_{change}$ and *all* situations. The situation assumed for these experiments was $SIT_{multplans}$. Three dense random (subject to the above-mentioned restrictions) FSAs were generated, and then the product was formed. The result was a product FSA satisfying the property. Operator $o_{change}$ was then applied, which consisted of a random (but points to a state instead of 0) change to a randomly chosen table entry in the FSA transition table for a random choice of one of the three agents. Finally, the product FSA was re-formed and reverification done.

The methodology for generating Table 3 was similar to that for Table 2, except that $o_{gen}$ was the learning operator and the situation was assumed to be $SIT_{1plan}$. In other words, the product FSA was formed, and then $o_{gen}$ applied to the *product* FSA of the three agents, the product was taken with the property FSA if needed for AT verification, and then reverification performed. Operator $o_{gen}$ consisted of choosing a random state $s_i$ and a random action $a_i$ for which $\delta(s_i, a_i) = s_k$, and choosing a random action $a_j$ for which $\delta(s_i, a_j) = 0$, and then setting $\delta(s_i, a_j) = s_k$.

Any column in Tables 2 or 3 labeled "sec" gives a mean, over 30 runs, of the cpu time of the algorithm. Columns labeled "spd" give the speedup over total, i.e., the cpu time of the incremental algorithm in that row divided by the cpu time of the corresponding total algorithm. For example, the "spd" entry for $Inc_{prod}$ in row 1 gives its cpu time divided by the cpu time of $Total_{prod}$ in row 7. Columns labeled "err" show the average number of verification errors over 30 runs. This is important to monitor because, for example, the cpu time is most strongly correlated with the number of states "visited" during dfs, and "visited2" during ndfs when AT verification is used. Every property error causes ndfs to be called with a nested search, which may be quite time-consuming. Also, it is important to note that we did not force any verification errors to occur. It was our objective to monitor cpu time under natural circumstances for evolving FSAs. When errors arose they were the natural result of applying a learning operator. The "err" columns are missing from Table 2 because the values are all 0, i.e., no errors occurred during the experiments due to applying $o_{change}$, although we have observed errors to occur with this operator not during the experiments. The lack of errors in the experiments resulted from the particular random FSAs that happened to be generated during the experiments. Errors are quite common with the specific $o_{gen}$ version of $o_{change}$, as can be seen in Table 3. Note that "N/A" is in the "err" column for anything other than a verification algorithm because "err" refers to verification errors.

The algorithms (rows) should be considered in triples "p," "v," and "b," or else as a single item "v+b." A "p" next to an algorithm name in Table 2 or 3 denotes it is a product algorithm, a "v" that it is a verification algorithm, and a "b" that it is the sum of the "p" and "v" entries, i.e., the time for *both* re-forming the product and reverifying. For example, $Inc_I$ (b) is considered to be an algorithm pair consisting of $Inc_{prod}$ (p) followed by $Total_I$ (v) (see rows 1-3 of Table 2). If no product needs to be formed, then the "b" version of the algorithm is identical to the "v" version, in which case there is only one row labeled "v+b."

Tables 4, 5, and 6 *re*-present a subset (cpu time only) of the data from Tables 2 and 3 in a format that facilitates some comparisons. In other words, Tables 4, 5, and 6 contain

no new data, only reformatted data from Tables 2 and 3. In Tables 4, 5, and 6, results are grouped by "p," "v," or "b."

Let us elaborate on one more interesting issue before listing our experimental hypotheses. Recall that we are using a First-Response property FSA and that this FSA checks only that the *first* trigger in every string is followed by a response. For our evolutionary paradigm (with dense FSA initialization) when using $Inc_{AT-NI}$, verifying a First-Response property is equivalent to verifying the full Response property. The false errors found by $Inc_{AT-NI}$ due to its incompleteness are in fact violations of the full Response property.[16] Therefore for $Inc_{AT-NI}$, First-Response FSAs are entirely adequate for reverification of full Response properties. Because we used the evolutionary paradigm in these experiments, and because $Inc_{AT-NI}$ found the same number of errors as $Total_{AT}$ (i.e., $Inc_{AT-NI}$ found no false errors), for the FSAs in these experiments testing First-Response properties was equivalent to testing full Response properties.

For our experiments, five hypotheses were tested:

H1: Algorithms tailored specifically for Invariance properties are faster than those for AT verification, because the latter are general-purpose (and the product algorithms include an additional FSA).

H2: The incremental algorithms are faster than the total algorithms for both product and reverification. This is expected to be true because they were tailored for learning.

H3: The "NI" versions of the incremental algorithms are faster than their counterparts, which do not find new initial states. This is expected because of the increase in streamlining.

H4: $Inc_{gen-I}$ and $Inc_{gen-R}$ are the fastest of all the algorithms, because they are tailored for a less generic learning operator (i.e., $o_{gen}$ rather than $o_{change}$), *plus* they are also tailored for one specific property type, and they sacrifice finding all errors.

H5: $Inc_{gen-I}$ and $Inc_{gen-R}$ will have the best scaleup properties. They will not take more time as FSA size increases. This latter expectation comes from the worst-case time complexity analysis.

Subsidiary issues we examine are the percentage of wrong predictions (for $Inc_{AT-NI}$ and $Inc_{gen-R}$, which are not complete algorithms), and the maximum observed speedup.

The results are the following (unless stated otherwise, look at the "sec" columns):

H1: To see the results, in Table 2 look at rows 1 through 9 and compare each row $r$ in this set with row $r+9$. In other words, compare row 1 with row 10, row 2 with row 11, and so on. Rows 1 through 9 are algorithms for Invariance properties, and rows 10 through 18 are algorithms for AT verification. In Table 3, rows 1 through 3 are algorithms for Invariance properties, and rows 5 through 10 are algorithms for AT verification. Compare row 2 with 7, and 3 with 10. (Rows 1 and 4 cannot be compared because row 4 has an algorithm tailored for Response properties.) Note that these comparisons are between a "v+b" and a "b." Since "v+b" means "v" or "b," this is a correct comparison. **These results show that H1 is mostly, but not completely, confirmed.** It is confirmed for all results in Table 3. On the other hand, the results are mixed for Table 2.

---

16. The reason is the following. Dense FSA initialization creates FSAs with no triggers. A learning operator is then applied. After learning, $Inc_{AT-NI}$ begins reverification at *every* state from which a new trigger could have been added by learning. Thus *every* trigger in the FSA will be checked to see if it is followed by a response. At every generation of our evolutionary learning paradigm, at most one learning operator is applied per FSA, and this is immediately followed by reverification and error resolution (if needed). Therefore every new trigger will be caught by $Inc_{AT-NI}$ and, if not followed by a response, the problem will be immediately resolved.

|   |   | 25-state FSAs | | 45-state FSAs | |
|---|---|---|---|---|---|
|   |   | sec | spd | sec | spd |
| 1 | $Inc_{prod}$ **p** | .000157 | .00497 | .000492 | .00255 |
| 2 | $Total_I$ **v** | .023798 | .95663 | .206406 | .97430 |
| 3 | $Inc_I$ **b** | .023955 | .07023 | .206898 | .51133 |
| 4 | $Inc_{prod-NI}$ **p** | .000206 | .00652 | .000617 | .00320 |
| 5 | $Inc_{I-NI}$ **v** | .000169 | .00680 | .000528 | .00320 |
| 6 | $Inc_{I-NI}$ **b** | .000375 | .00110 | .001762 | .00435 |
| 7 | $Total_{prod}$ **p** | .031594 | 1.0 | .192774 | 1.0 |
| 8 | $Total_I$ **v** | .024877 | 1.0 | .211851 | 1.0 |
| 9 | $Total_I$ **b** | .340817 | 1.0 | .404625 | 1.0 |
| 10 | $Inc_{prod}$ **p** | .000493 | .00507 | .001521 | .00259 |
| 11 | $Total_{AT}$ **v** | .021103 | .98903 | .177665 | .96869 |
| 12 | $Inc_{AT}$ **b** | .024798 | .20022 | .180707 | .23441 |
| 13 | $Inc_{prod-NI}$ **p** | .000574 | .00590 | .001786 | .00304 |
| 14 | $Inc_{AT-NI}$ **v** | .009011 | .37450 | .090824 | .49520 |
| 15 | $Inc_{AT-NI}$ **b** | .009585 | .07900 | .092824 | .12013 |
| 16 | $Total_{prod}$ **p** | .097262 | 1.0 | .587496 | 1.0 |
| 17 | $Total_{AT}$ **v** | .024062 | 1.0 | .183409 | 1.0 |
| 18 | $Total_{AT}$ **b** | .121324 | 1.0 | .770905 | 1.0 |

Table 2: Average performance over 30 runs (5 properties, 6 runs each) with operator $o_{change}$ and dense FSAs. Rows 1 through 9 are for reverification of Invariance properties and rows 10 through 18 are for AT reverification of Response properties.

|   |   | 25-state FSAs | | | 45-state FSAs | | |
|---|---|---|---|---|---|---|---|
|   |   | sec | spd | err | sec | spd | err |
| 1 | $Inc_{gen-I}$ **v+b** | .000001 | 4.25e-5 | .20 | .000002 | 9.75e-6 | .07 |
| 2 | $Inc_{I-NI}$ **v+b** | .000002 | 8.51e-5 | .20 | .000003 | 1.46e-5 | .07 |
| 3 | $Total_I$ **v+b** | .023500 | 1.0 | .20 | .205082 | 1.0 | .07 |
| 4 | $Inc_{gen-R}$ **v+b** | .000007 | 7.23e-8 | .73 | .000006 | 2.09e-9 | .73 |
| 5 | $Inc_{prod-NI}$ **p** | .000006 | 5.22e-5 | N/A | .000006 | 8.51e-6 | N/A |
| 6 | $Inc_{AT-NI}$ **v** | 94.660700 | .98099 | 3569.33 | 2423.550000 | .84442 | 12553.40 |
| 7 | $Inc_{AT-NI}$ **b** | 94.660706 | .97982 | N/A | 2423.550006 | .84421 | N/A |
| 8 | $Total_{prod}$ **p** | .114825 | 1.0 | N/A | .704934 | 1.0 | N/A |
| 9 | $Total_{AT}$ **v** | 96.495400 | 1.0 | 3569.33 | 2870.080000 | 1.0 | 12553.40 |
| 10 | $Total_{AT}$ **b** | 96.610225 | 1.0 | N/A | 2870.784934 | 1.0 | N/A |

Table 3: Average performance over 30 runs (5 properties, 6 runs each) with operator $o_{gen}$ and dense FSAs. Rows 1 through 3 are for reverification of Invariance properties and rows 4 through 10 are for reverification of Response properties.

|   |   | **25-state FSAs** | **45-state FSAs** |
|---|---|---|---|
| **1** | $Inc_{prod}$ **p** | .000157 | .000492 |
| **2** | $Inc_{prod-NI}$ **p** | .000206 | .000617 |
| **3** | $Total_{prod}$ **p** | .031594 | .192774 |
| **4** | $Total_I$ **v** | .023798 | .206406 |
| **5** | $Inc_{I-NI}$ **v** | .000169 | .000528 |
| **6** | $Total_I$ **v** | .024877 | .211851 |
| **7** | $Inc_I$ **b** | .023955 | .206898 |
| **8** | $Inc_{I-NI}$ **b** | .000375 | .001762 |
| **9** | $Total_I$ **b** | .340817 | .404625 |

Table 4: Average cpu time (in seconds) over 30 runs with operator $o_{change}$ and five Invariance properties. This table is a duplication of some of the material in Table 2.

|   |   | **25-state FSAs** | **45-state FSAs** |
|---|---|---|---|
| **1** | $Inc_{prod}$ **p** | .000493 | .001521 |
| **2** | $Inc_{prod-NI}$ **p** | .000574 | .001786 |
| **3** | $Total_{prod}$ **p** | .097262 | .587496 |
| **4** | $Total_{AT}$ **v** | .021103 | .177665 |
| **5** | $Inc_{AT-NI}$ **v** | .009011 | .090824 |
| **6** | $Total_{AT}$ **v** | .024062 | .183409 |
| **7** | $Inc_{AT}$ **b** | .024798 | .180707 |
| **8** | $Inc_{AT-NI}$ **b** | .009585 | .092824 |
| **9** | $Total_{AT}$ **b** | .121324 | .770905 |

Table 5: Average cpu time (in seconds) over 30 runs with operator $o_{change}$ and five Response properties. This table is a duplication of some of the material in Table 2.

|   |   | **25-state FSAs** | **45-state FSAs** |
|---|---|---|---|
| **1** | $Inc_{gen-R}$ **p** | 0 | 0 |
| **2** | $Inc_{prod}$ **p** | .000006 | .000006 |
| **3** | $Total_{prod}$ **p** | .114825 | .704934 |
| **4** | $Inc_{gen-R}$ **v** | .000007 | .000006 |
| **5** | $Inc_{AT}$ **v** | 94.660700 | 2423.550000 |
| **6** | $Total_{AT}$ **v** | 96.495400 | 2870.080000 |
| **7** | $Inc_{gen-R}$ **b** | .000007 | .000006 |
| **8** | $Inc_{AT}$ **b** | 94.660706 | 2423.550006 |
| **9** | $Total_{AT}$ **b** | 96.610225 | 2870.784934 |

Table 6: Average cpu time (in seconds) over 30 runs with operator $o_{gen}$ and five Response properties. This table is a duplication of some of the material in Table 3.

H2: The easiest way to compare is to examine Tables 4, 5, and 6. In these cases the comparison is between the first two rows labeled "p" (or "v" or "b") versus the third row of that same label. The reason for making these comparisons is that the first two rows of a given label correspond to an incremental algorithm (except for row 4 of Tables 4 and 5) and the third row of a given label corresponds to a total algorithm. Alternatively, one could examine Tables 2 and 3. In Table 2, rows 1 through 6 (other than 2) and 10 through 15 (other than 11) are incremental algorithms, and rows 2, 11, 7 through 9, and 16 through 18 are total reverification algorithms. The appropriate comparisons are between rows 1 and 7, 4 and 7, 5 and 8, 3 and 9, 6 and 9, 10 and 16, 13 and 16, 14 and 17, 12 and 18, and 15 and 18. In Table 3, rows 1, 2, and 4 through 7 are incremental algorithms, and rows 3 and 8 through 10 are total. The appropriate comparisons are between rows 1 and 3, 2 and 3, 4 and 10, 5 and 8, 6 and 9, and 7 and 10. **All results confirm H2.** The statistical significance of the comparisons in Tables 2 and 3 were tested. Using an exact Wilcoxon rank-sum test, all comparisons relevant to hypothesis H2 in Table 2 are statistically significant ($p < 0.01$ and, in most cases, $p < 0.0001$). In Table 3, however, the differences between $Inc_{AT-NI}$ and $Total_{AT}$ (both the (v) and (b) versions) are not statistically significant at the $p < 0.01$ level. All other comparisons in Table 3 are significant at the $p < 0.01$ level.

H3: This hypothesis does not apply to the algorithms for re-forming the product FSA because, obviously, it will require more time to get the new initial states for the "NI" versions. We wish to test the *overall* time savings of the "NI" versions, so we concentrate on the rows labeled "b." The relevant comparisons are row 7 versus 8 in Table 4 and row 7 versus 8 in Table 5. (Alternatively, one could compare row 3 versus 6, and row 12 versus 15 in Table 2.) Each of these comparisons is between an "NI" version and a counterpart version of the algorithm that is the same as the "NI" version except that it does not find new initial states. Tables 3 and 6 are not relevant because they only have the "NI" versions but not their counterparts. (We only saw the need to make one comparison between all "NI" versions and their counterparts, which is reflected in Table 2.) **All results confirm hypothesis H3.** After testing the statistical significance, it is found that the results are significant ($p < 0.01$).

H4: To determine H4 requires considering Table 3 but not Table 2. This is because we only need to compare algorithms for which $o_{gen}$ has been applied. Compare row 1 versus 2, 1 versus 3, 4 versus 7, and 4 versus 10 to see the results. All results show $Inc_{gen-I}$ (row 1) and $Inc_{gen-R}$ (row 4) to be at least as fast as the other algorithms. **Therefore H4 is confirmed.** In all cases other than $Inc_{gen-I}$ (row 1) versus $Inc_{I-NI}$ (row 2), there is a noticeable speedup. In most cases, the speedup is quite dramatic. All noticeable speedups are statistically significant ($p < 0.0001$).

H5: To test H5, compare the first "spd" column (for 25-state FSAs) with the second column with this label (for 45-state FSAs). A more desirable scaleup shows a lower value for "spd" as the size of the FSA increases. It implies that the ratio of the cpu time of the incremental algorithm to the cpu time of the total algorithm decreases more (or increases less) as the FSA size increases. One should make this two-column comparison for rows 1 through 6 (but not 2) and 10 through 15 (but not 11) of Table 2, and rows 1 and 2, and 4 through 7 of Table 3 because these are all the incremental algorithms. (We don't care about the total algorithms because "spd" is, by definition, always 1.0 for them.)[17] If one considers the results of algorithms appearing in both tables (e.g., $Inc_{I-NI}$ shows different scaleup properties in the two tables, but we need to consider both sets of results), then clearly $Inc_{gen-I}$ (row 1) and $Inc_{gen-R}$ (row 4) in Table 3 show the best scaleup of all the incremental algorithms. **H5 is confirmed.** It is apparent from the "sec" columns that the time complexity of these two algorithms does not increase (other than minor fluctuations) as FSA size increases (see Table 3).

A couple of subsidiary issues are now addressed. For one, recall that $Inc_{AT-NI}$ and $Inc_{gen-R}$ are not complete. Therefore, it is relevant to consider the percentage of incorrect predictions (i.e., false errors) they made. $Inc_{AT-NI}$ made none. For the results in Table 3, 33% of $Inc_{gen-R}$'s predictions were wrong (i.e., false errors) for the size 25 FSAs, and 50% were wrong for the size 45 FSAs.

Finally, consider the maximum observable speedup. $Inc_{gen-R}$ shows a $\frac{1}{2}$-*billion-fold speedup* over $Total_{AT}$ on size 45 FSA problems (averaged over 30 runs)! This alleviates much of the concern about $Inc_{gen-R}$'s false error rate. For example, given the rapid reverification time of $Inc_{gen-R}$, an agent could use it to reverify a long sequence of learning operators culminating in one that satisfies the property in considerably less time than it takes $Total_{AT}$ to reverify one learning operator.

We conclude this section by summarizing, in Table 7, the fastest algorithm (based on our results) for every operator, situation, and property type. In Table 7, it is assumed that a First-Response FSA is used for AT verification of Response properties. Operator $o_{add-action}$ is omitted from this table because it is not clear at this time whether it would be faster to apply total reverification or perform multiple applications of the incremental algorithm (one for each primitive operator application). Section 8 considers an alternative solution as future work. In Table 7, "None" means no reverification is required, i.e., the learning operator is a priori guaranteed to be an SML for this situation and property class.

## 7. Related Work

There has been a great deal of recent research on model checking, and even on model checking of distributed systems (Holzmann, 1991). Nevertheless, there is very little in the literature about model checking applied to systems that change. Two notable exceptions are the research of Sokolsky and Smolka (1994) on incremental reverification and that of

---

17. If "spd" $\neq 1.0$ for a total algorithm, this is due to the statistical variation in run time.

| | $SIT_{1agent/1plan}$ and Invariance | $SIT_{1agent/1plan}$ and Response | $SIT_{multplans}$ and Invariance | $SIT_{multplans}$ and Response |
|---|---|---|---|---|
| $o_{change}$ | $Inc_{I-NI}$ | $Inc_{AT-NI}$ | $Inc_{I-NI}$ | $Inc_{AT-NI}$ |
| $o_{delete}$ | $None$ | $None$ | $None$ | $None$ |
| $o_{spec}$ | $None$ | $None$ | $None$ | $None$ |
| $o_{add}$ | $Inc_{I-NI}$ | $Inc_{AT-NI}$ | $Inc_{I-NI}$ | $Inc_{AT-NI}$ |
| $o_{gen}$ | $Inc_{gen-I}$ or $Inc_{I-NI}$ | $Inc_{gen-R}$ | $Inc_{I-NI}$ | $Inc_{AT-NI}$ |
| $o_{delete \vee spec}$ | $None$ | $None$ | $None$ | $None$ |
| $o_{delete-action}$ | $None$ | $None$ | $None$ | $None$ |
| $o_{add \vee gen}$ | $Inc_{I-NI}$ | $Inc_{AT-NI}$ | $Inc_{I-NI}$ | $Inc_{AT-NI}$ |
| $o_{move}$ | $Inc_{I-NI}$ | $Inc_{AT-NI}$ | $Inc_{I-NI}$ | $Inc_{AT-NI}$ |
| $o_{delete+add}$ | $Inc_{I-NI}$ | $Inc_{AT-NI}$ | $Inc_{I-NI}$ | $Inc_{AT-NI}$ |
| $o_{spec+add}$ | $Inc_{I-NI}$ | $Inc_{AT-NI}$ | $Inc_{I-NI}$ | $Inc_{AT-NI}$ |
| $o_{delete+gen}$ | $None$ | $Inc_{gen-R}$ | $Inc_{I-NI}$ | $Inc_{AT-NI}$ |
| $o_{spec+gen}$ | $None$ | $Inc_{gen-R}$ | $Inc_{I-NI}$ | $Inc_{AT-NI}$ |
| $o_{stay}$ | $None$ | $Inc_{AT-NI}$ | $Inc_{I-NI}$ | $Inc_{AT-NI}$ |

Table 7: Learning operators with the fastest reverification method.

Sekar et al. (1994). Both of these papers are about reverification of software after user edits rather than adaptive agents. Nevertheless the work is related. Sokolsky and Smolka use the modal $\mu$-calculus to express Invariance and Liveness properties. They present an incremental version of a model checker that does block-by-block global computations of fixed points, rather than AT or property-specific model checking as we do. The learning operators assumed by their algorithm are edge deletions/additions on a representation similar to FSAs called LTS (but unlike our multiagent work, they assume a single LTS). The worst-case time complexity of their algorithm is the same as that of total reverification, although their empirical results are good. Note that we have a priori results for edge deletion. However we do not have an incremental algorithm specifically tailored for edge addition (for multiple agents and AT or property-specific model checking); thus this may be a fruitful direction for future research. Sekar et al.'s approach consists of converting rule sets to FSAs, then generating and testing functions that map from the post- to the prelearning FSA and property. If the desired function can be found, they apply a theorem from Kurshan (1994), which guarantees that the learning is "safe." Although no complexity results are provided, the generate-and-test approach that they describe appears to be computationally expensive. In contrast to Sekar et al., we have proofs and empirical evidence that our methods are efficient and, in some cases, that they are substantially more efficient than total reverification from scratch.

There is also related research in the field of classical planning. In particular, Weld and Etzioni (1994) have a method to incrementally test an agent's plan to decide whether to add new actions to the plan. Actions are added only when their effects do not violate a certain type of Invariance property. Their method has some similarities with our $Inc_{gen-I}$ algorithm. One difference is that our method is for reactive rather than projective plans.

Another is that our verification method is expressed using the formal foundations in the model checking literature.

As mentioned in the introduction of this paper, FSAs have been shown to be effective representations of reactive agent plans/strategies (Burkhard, 1993; Kabanza, 1995; Carmel & Markovitch, 1996; Fogel, 1996). FSA plans have been used both for multiagent competition and coordination. For example, Fogel's (1996) co-evolving FSA agents for competitive game playing were mentioned above. A similarity with our work is that Fogel assumes agents' plans are expressed as $\omega$-automata. Nevertheless, Fogel never discusses verification of these plans. Goldman and Rosenschein (1994) present a method for multiagent coordination that assumes FSA plans. Multiple agents cooperate by taking actions to favorably alter their environment. The cooperation strategy is implemented by a plan developer who manually edits the FSAs. The relationship to the work here is that they present FSA transformations that ensure multiagent coordination. Likewise, in our research, a learning operator that is a priori guaranteed "safe" for some multiagent coordination property transforms the FSA while ensuring coordination. Although both their method and ours guarantee this coordination, their solution is manual whereas ours is entirely automated.

Some of the more recent research on agent coordination applies formal verification methods. For example, Lee and Durfee (1997) model their agents' semantics with a formalism similar to Petri nets (rather than FSAs). They verify synchronization (Invariance) properties, which prevent deadlock, using model checking. Furthermore, Lee and Durfee suggest recovery from failed verification using two methods: concept learning, and a method analogous to that used by Ramadge and Wonham (1989). Burkhard (1993) and Kabanza (1995) assume agent plans are represented as $\omega$-automata, and they address issues of model checking temporal logic properties of the joint (multiagent) plans. Thus there is a growing precedent for addressing multiagent coordination by expressing plans as $\omega$-automata and verifying them with model checking. Our work builds on this precedent, and also extends it, because none of this previous research addresses efficient *re*verification for agents that learn.

Finally, there are alternative methods for constraining the behavior of agents, which are complementary to reverification and self-repair. For example, Shoham and Tennenholtz (1995) design agents that obey social laws, e.g., safety conventions, by restricting the agents' actions. Nevertheless, the plan designer may not be able to anticipate and engineer all laws into the agents beforehand, especially if the agents have to adapt. One solution is to use laws that allow maximum flexibility (Fitoussi & Tennenholtz, 1998). However this solution does not allow for certain changes in the plan, such as the addition or deletion of actions. An appealing alternative would be to couple initial engineering of social laws with efficient reverification after learning.

A method for ensuring physically bounded behavior of agents is "artificial physics" (Spears & Gordon, 1999). With artificial physics, multiagent behavior is restricted by artificial forces between the agents. Nevertheless, when encountering severe unanticipated circumstances, artificial physics needs to be complemented with reverification and "steering" for self-repair (Gordon et al., 1999).

# 8. Summary and Future Work

Agent technology is growing rapidly in popularity. To handle real-world domains and interactions with people, agents must be adaptable, predictable, *and* rapidly responsive. An approach to resolving these potentially conflicting requirements is presented here. In summary, we have shown that certain machine learning operators are a priori (with no run-time reverification) safe to perform. In other words, when certain desirable properties hold prior to learning, they are guaranteed to hold post-learning. The property classes considered here are Invariance and Response. Learning operators $o_{delete}$, $o_{spec}$, $o_{delete \vee spec}$, and $o_{delete-action}$ were found to preserve properties in either of these classes. For $SIT_{1agent}$ and $SIT_{1plan}$, where there is a single (multi)agent FSA plan, $o_{delete+gen}$, $o_{spec+gen}$ and $o_{stay}$ were found to preserve Invariance properties. All of the a priori results are independent of the size of the FSA and are therefore applicable to any FSA that has been model checked originally.

We then discussed transformations of learning operators and their corresponding a priori results to a product plan. This addresses $SIT_{multiplans}$, where multiple agents each have their own plan but the multiagent plan must be re-formed and reverified to determine whether multiagent properties are preserved. It was discovered that only $o_{delete}$, $o_{spec}$, $o_{delete \vee spec}$, and $o_{delete-action}$ preserve their a priori results for this situation.

Finally, we presented novel incremental reverification algorithms for all cases in which the a priori results are negative. It was shown in both theoretical and empirical comparisons that these algorithms can substantially improve the time complexity of reverification over total reverification from scratch. Empirical results showed as much as a $\frac{1}{2}$-billion-fold speedup. These are initial results, but continued research along these lines will likely be applicable to a wide range of important problems, including a variety of agent domains as well as more general software applications.

When learning is required, we suggest that the a priori results should be consulted first. If no positive results (i.e., the learning operator is an SML) exist, then incremental reverification proceeds.

To test our overall framework, we have implemented the rovers example of this paper as co-evolving agents assuming $SIT_{multiplans}$, i.e., multiple agents each with its own plan. By using the a priori results and incremental algorithms, we achieved significant speedups. We have also developed a more sophisticated application that uses reverification during evolution. Two agents compete in a board game, and one of the agents evolves its strategy to improve it. The key lesson that has been learned from this implementation is that although the types of FSAs and learning operators are slightly different from those presented in this paper, and the property is quite different (it is a check for a certain type of cyclic behavior on the board), initial experiences show that the methodology and basic results here could potentially be easily extended to a variety of multiagent applications.

Future work will focus primarily on extending the a priori results to other learning operators/methods and property classes, developing other incremental reverification algorithms, and exploring plan repair to recover from reverification failures. One way in which the a priori results might be extended is by discovering when learning operators will make a property true, even if it was not true before learning.

A question that was not addressed here is whether the incremental methods are useful if multiple machine learning operators are applied in batch (e.g., as one might wish to do with operator $o_{add-action}$). In the future we would like to explore how to handle this situation – is it more efficient to treat the operators as having been done one-at-a-time and use incremental reverification for each? Or is total reverification from scratch preferable? Or, better yet, can we develop efficient incremental algorithms for *sets* of learning operators?

Plan repair was not discussed in this paper and is an important future direction. The research of De Raedt and Bruynooghe (1994), which uses counterexamples to guide the revision of theories subject to *integrity constraints*, may provide some ideas. There are also plan repair methods in the classical planning literature that might be relevant to our approach (Joslin & Pollack, 1994; Weld & Etzioni, 1994). It would be interesting to compare the time to repair plans versus trying another learning operator and reverifying.

A limitation of our approach is that it does not handle stochastic plans or properties with time limits, e.g., a Response property for which the response must occur within a specified time after the trigger. We would like to extend this research to stochastic FSAs (Tzeng, 1992) and timed FSAs/properties (Alur & Dill, 1994; Kabanza, 1995), as well as other common agent representations besides FSAs. Another direction for future work would be to extend our results to symbolic model checking, which uses binary decision diagrams (BDDs) so that the full state space need not be explicitly explored during model checking (Burch et al., 1994). In some cases, symbolic model checking can produce dramatic speedup. However, none of the current research on symbolic model checking addresses adaptive systems.

Additionally, the ideas here are applicable to some of the FSA-based control theory work. For example, Ramadge and Wonham (1989) assume FSA representations for both the plant (which is assumed to be a discrete-event system) and the supervisor (which controls the actions of the plant). We are currently applying some of the principles of efficient reverification to change the supervisor in response to changes in the plant in a manner that preserves properties (Gordon & Kiriakidis, 2000).

Finally, future work should focus on studying how to operationalize Asimov's Laws for intelligent agents. What sorts of properties best express these laws? Weld and Etzioni (1994) provide some initial suggestions, but much more remains to be done.

## Acknowledgments

This research is supported by the Office of Naval Research (N0001499WR20010) in conjunction with the "Semantic Consistency" MURI. I am grateful to William Spears, Joseph Gordon, Stan Sadin, Chitoor Srinivasan, Ramesh Bharadwaj, Dan Hoey, and the anonymous reviewers for useful suggestions and advice. The presentation of the material in this paper was enormously improved thanks to William Spears' suggestions.

## Appendix A. Glossary of Notation

| | |
|---|---|
| $\models$ | Models (satisfies) |
| *model checking* | A verification method entailing brute-force search |
| AT | Automata-theoretic model checking |
| $SIT_{1agent}$ | Single agent situation |
| $SIT_{1plan}$ | Multiagent situation where each agent uses a multiagent plan |
| $SIT_{multplans}$ | Multiagent situation where each agent uses an individual plan |
| $FSA$ | Finite-state automaton |
| $V(S)$ | The set of states (vertices) of FSA $S$ |
| $E(S)$ | The set of state-to-state transitions (edges) of FSA $S$ |
| *transition condition* | Logical description of the set of actions enabling a transition |
| $\mathcal{K}$ | A Boolean algebra |
| $\preceq$ | Boolean algebra partial order; $x \preceq y$ iff $x \wedge y = x$ |
| $M_{\mathcal{K}}(S)$ | The matrix of transition conditions of FSA $S$ |
| $M_{\mathcal{K}}(v_i, v_j)$ | Transition condition associated with edge $(v_i, v_j)$ |
| $I(S)$ | The set of initial states of FSA $S$ |
| *atoms* | Primitive elements of a Boolean algebra; atoms are actions |
| *string* | Sequence of actions (atoms) |
| $\mathcal{L}(S)$ | The language of (set of strings accepted by) FSA $S$ |
| $\omega$-automaton | An FSA that accepts infinite-length strings |
| *run* | The sequence of FSA vertices visited by a string |
| *accepting run* | The run of a string in the FSA language |
| *acceptance criterion* | A requirement of accepting runs of an FSA |
| $\otimes$ | The tensor (synchronous) product of FSAs |
| *complete FSA* | Specifies a transition for every possible action |
| *deterministic FSA* | The choice of action uniquely determines the next state |
| *path* | Sequence of vertices connected by edges |
| *cycle* | A path with start and end vertices identical |
| *c-state* | Computational state; an action occurring in a computation |
| *accessible from* | There exists a path from |
| $\square$ | Temporal logic "invariant" |
| $\diamond$ | Temporal logic "eventually" |
| *Invariance property* | $\square \neg p$, i.e., "Invariant not $p$" |
| *Response property* | $\square(p \rightarrow \diamond q)$, i.e., "Every $p$ is eventually followed by $q$" |
| *First-Response property* | The first $p$ (trigger) is followed by a $q$ (response) |
| $B(S)$ | The set of "bad" (to be avoided) states of FSA $S$ |
| $\uparrow$ | Can increase accessibility |
| $\not\uparrow$ | Cannot increase accessibility |
| $\downarrow$ | Can decrease accessibility |
| $\not\downarrow$ | Cannot decrease accessibility |
| $SML$ | Safe machine learning operator, i.e., preserves properties |
| *sound algorithm* | One that is correct when it states that $S \models P$ |
| *complete algorithm* | One that is correct when it states that $S \not\models P$ |
| $\delta$ | The FSA transition function |

## Appendix B. Temporal logic properties

This appendix, which is based on Manna and Pnueli (1991), formally defines Invariance and Response properties in temporal logic. We begin by defining the basic temporal operator $\mathcal{U}$ (Until). We assume a string $(x_0, ...)$ of c-states of FSA $S$, where $0 \leq i, j, k$. Then for c-state formulae $p$ and $q$, we define Until as $x_j \models p \, \mathcal{U} \, q \Leftrightarrow$ for some $k \geq j$, $x_k \models q$, and for every $i$ such that $j \leq i < k$, $x_i \models p$.

Invariance properties are defined in terms of Eventually properties, so we define Eventually first. For c-state formula $p$ and FSA $S$, we define property $P = \Diamond p$ ("Eventually $p$") as a property that is true (false) for a string if it is true (false) at the initial c-state $x_0$ of the string. Formally, if $\mathbf{x} = (x_0, ...)$ is a string of FSA $S$, then $\mathbf{x} \models \Diamond p \Leftrightarrow x_0 \models true \, \mathcal{U} \, p$, i.e., "eventually $p$." A property $P = \Box \neg p$ ("Invariant not $p$") is defined as $\mathbf{x} \models \Box \neg p \Leftrightarrow \mathbf{x} \models \neg \Diamond p$, i.e., "never $p$." Finally, a Response formula is of the form $\Box(p \rightarrow \Diamond q)$, where $p$ is called the "trigger" and $q$ the "response." A Response formula states that every trigger is eventually followed by a response.

## Appendix C. Properties for Experiments

The following five Invariance properties were used in the test suite:

$\Box$ ($\neg$(I-deliver $\wedge$ L-transmit))

$\Box$ ($\neg$(I-deliver $\wedge$ L-pause))

$\Box$ ($\neg$(F-collect $\wedge$ I-deliver))

$\Box$ ($\neg$(F-collect $\wedge$ I-deliver $\wedge$ L-receive))

$\Box$ ($\neg$(F-deliver $\wedge$ I-receive $\wedge$ L-pause))

The following five Response properties were used in the test suite:

$\Box$ (F-deliver $\rightarrow \Diamond$ L-receive)

$\Box$ (F-deliver $\rightarrow \Diamond$ I-receive)

$\Box$ (F-collect $\rightarrow \Diamond$ L-transmit)

$\Box$ ((F-collect $\wedge$ I-deliver) $\rightarrow \Diamond$ L-receive)

$\Box$ (F-deliver $\rightarrow \Diamond$ (I-receive $\wedge$ L-receive))

## References

Alur, R., & Dill, D. (1994). A theory of timed automata. *Theoretical Computer Science*, *126*, 183–235.

Asimov, I. (1950). *I, Robot*. Greenwich, CT: Fawcett Publications, Inc.

Bavel, Z. (1983). *Introduction to the Theory of Automata*. Reston, VA: Prentice-Hall.

Büchi, J. (1962). On a decision method in restricted second-order arithmetic. In *Methodology and Philosophy of Science, Proceedings of the Stanford International Congress*, pp. 1–11. Stanford, CA: Stanford University Press.

Burch, J., Clarke, E., Long, D., McMillan, K., & Dill, D. (1994). Symbolic model checking for sequential circuit verification. *IEEE Transactions on Computer-Aided Design of Integrated Circuits and Systems, 13(4)*, 401–424.

Burkhard, H. (1993). Liveness and fairness properties in multi-agent systems. In *Proceedings of the Thirteenth International Joint Conference on Artificial Intelligence (IJCAI)*, pp. 325–330. Chambery, France.

Carmel, D., & Markovitch, S. (1996). Learning models of intelligent agents. In *Proceedings of the Thirteenth National Conference on Artificial Intelligence (AAAI)*, pp. 62–67. Portland, OR.

Clarke, E., & Wing, J. (1997). Formal methods: State of the art and future directions. *ACM Computing Surveys, 28(4)*, 626–643.

Courcoubetis, C., Vardi, M., Wolper, P., & Yannakakis, M. (1992). Memory-efficient algorithms for the verification of temporal properties. *Formal Methods in System Design, 1*, 257–288.

De Raedt, L., & Bruynooghe, M. (1994). Interactive theory revision. In Michalski, R., & Tecuci, G. (Eds.), *Machine Learning IV*, pp. 239–264. San Mateo, CA: Morgan Kaufmann.

Dean, T., & Wellman, M. (1991). *Planning and Control*. San Mateo, CA: Morgan Kaufmann.

Elseaidy, W., Cleaveland, R., & Baugh, J. (1994). Verifying an intelligent structure control system: A case study. In *Proceedings of the Real-Time Systems Symposium*, pp. 271–275. San Juan, Puerto Rico.

Fitoussi, D., & Tennenholtz, M. (1998). Minimal social laws. In *Proceedings of the Fifteenth National Conference on Artificial Intelligence*, pp. 26–31. Madison, WI.

Fogel, D. (1996). On the relationship between duration of an encounter and the evolution of cooperation in the iterated prisoner's dilemma. *Evolutionary Computation, 3(3)*, 349–363.

Goldman, S., & Rosenschein, J. (1994). Emergent coordination through the use of cooperative state-changing rules. In *Proceedings of the Twelfth National Conference on Artificial Intelligence*, pp. 408–413. Seattle, WA.

Gordon, D. (1998). Well-behaved borgs, bolos, and berserkers. In *Proceedings of the Fifteenth International Conference on Machine Learning (ICML)*, pp. 224–232. Madison, WI.

Gordon, D. (1999). Re-verification of adaptive agents' plans. Tech. rep., Navy Center for Applied Research in Artificial Intelligence.

Gordon, D., & Kiriakidis, K. (2000). Adaptive supervisory control of interconnected discrete event systems. In *Proceedings of the International Conference on Control Applications (ICCA)*, pp. 50–56. Anchorage, AK.

Gordon, D., Spears, W., Sokolsky, O., & Lee, I. (1999). Distributed spatial control, global monitoring and steering of mobile physical agents. In *Proceedings of the IEEE International Conference on Information, Intelligence, and Systems (ICIIS)*, pp. 681–688. Washington, D.C.

Grefenstette, J., & Ramsey, C. (1992). An approach to anytime learning. In *Proceedings of Ninth International Workshop on Machine Learning*, pp. 189–195. Aberdeen, Scotland.

Heitmeyer, C., Kirby, J., Labaw, B., Archer, M., & Bharadwaj, R. (1998). Using abstraction and model checking to detect safety violations in requirements specifications. *IEEE Transactions on Software Engineering*, *24(11)*, 927–948.

Holzmann, G., Peled, D., & Yannakakis, M. (1996). On nested depth-first search. In *Proceedings of the Second Spin Workshop*, pp. 81–89. Rutgers, NJ.

Holzmann, G. J. (1991). *Design and Validation of Computer Protocols*. NJ: Prentice-Hall.

Joslin, D., & Pollack, M. (1994). Least-cost flaw repair: A plan refinement strategy for partial-order planning. In *Proceedings of the Twelfth International Conference on Artificial Intelligence*, pp. 1004–1009. Seattle, WA.

Kabanza, F. (1995). Synchronizing multiagent plans using temporal logic specifications. In *Proceedings of the First International Conference on Multiagent Systems (ICMAS)*, pp. 217–224. San Francisco, CA.

Kurshan, R. (1994). *Computer Aided Verification of Coordinating Processes*. Princeton, NJ: Princeton University Press.

Lee, J., & Durfee, E. (1997). On explicit plan languages for coordinating multiagent plan execution. In *Proceedings of the Fourth International Workshop on Agent Theories, Architectures, and Languages (ATAL)*, pp. 113–126. Providence, RI.

Manna, Z., & Pnueli, A. (1991). Completing the temporal picture. *Theoretical Computer Science*, *83(1)*, 97–130.

Michalski, R. (1983). A theory and methodology of inductive learning. In Michalski, R., Carbonell, J., & Mitchell, T. (Eds.), *Machine Learning I*, pp. 83–134. Palo Alto, CA: Tioga.

Mitchell, T. (1978). *Version Space: An Approach to Concept Learning*. Ph.D. thesis, Stanford University.

Nilsson, N. (1980). *Principles of Artificial Intelligence*. Palo Alto, CA: Tioga.

Potter, M. (1997). *The Design and Analysis of a Computational Model of Cooperative Coevolution*. Ph.D. thesis, George Mason University.

Ramadge, P., & Wonham, W. (1989). The control of discrete event systems. *Proceedings of the IEEE*, *1*, 81–98.

Sekar, R., Lin, Y.-J., & Ramakrishnan, C. (1994). Modeling techniques for evolving distributed applications. In *Proceedings of Formal Description Techniques (FORTE)*, pp. 22–29. Berne, Switzerland.

Shoham, Y., & Tennenholz, M. (1995). On social laws for artificial agent societies: Off-line design. *Artificial Intelligence, 73(1-2)*, 231–252.

Sikorski, R. (1969). *Boolean Algebras*. New York, NY: Springer-Verlag.

Sokolsky, O., & Smolka, S. (1994). Incremental model checking in the modal mu-calculus. In *Proceedings of Computer-Aided Verification (CAV)*, pp. 351–363. Stanford, CA.

Spears, W., & Gordon, D. (1999). Using artificial physics to control agents. In *Proceedings of the IEEE International Conference on Information, Intelligence, and Systems*, pp. 281–288. Washington, D.C.

Tennenholtz, M., & Moses, Y. (1989). On cooperation in a multi-entity model. In *Proceedings of the Eleventh International Joint Conference on Artificial Intelligence*, pp. 918–923.

Tzeng, W. (1992). Learning probabilistic automata and markov chains via queries. *Machine Learning, 8*, 151–166.

Vardi, M., & Wolper, P. (1986). An automata-theoretic approach to automatic program verification. In *Proceedings of the First Annual Symposium on Logic in Computer Science (LICS)*, pp. 332–345. Cambridge, MA.

Weld, D., & Etzioni, O. (1994). The first law of robotics. In *Proceedings of the Twelfth National Conference on Artificial Intelligence*, pp. 1042–1047. Seattle, WA.